\pdfoutput=1

\documentclass[11pt,table]{article}
\usepackage{silence}
\usepackage[final]{acl}

\usepackage{times}
\usepackage{latexsym}
\usepackage{amsmath, amssymb, amsthm}

\theoremstyle{plain}

\usepackage[T1]{fontenc}

\usepackage[utf8]{inputenc}

\usepackage{microtype}

\usepackage{inconsolata}

\usepackage{graphicx}
\usepackage{booktabs}   
\usepackage{multirow}   
\usepackage{stfloats}   
\usepackage{amsmath} 
\usepackage{colortbl}
\usepackage{array}
\usepackage{tabularx} 
\usepackage{float}
\usepackage{placeins}  
\usepackage[T1]{fontenc}
\usepackage{siunitx} 
\usepackage{makecell}   
\usepackage{arydshln} 
\usepackage{wrapfig}
\usepackage{ragged2e}

\newcommand{\checkedstar}{\checkmark\!*} 

\definecolor{softgreen}{RGB}{204, 255, 204} 
\definecolor{lightgreen}{RGB}{220, 255, 220} 
\definecolor{verylightgreen}{RGB}{235, 255, 235} 
\definecolor{softorange}{RGB}{255, 224, 204} 
\definecolor{lightgray}{RGB}{230, 230, 230} 
\definecolor{softblue}{RGB}{204, 229, 255}  
\definecolor{lightpink}{RGB}{255, 204, 229}  
\definecolor{lavender}{RGB}{230, 230, 255}  
\definecolor{peach}{RGB}{255, 204, 178}     
\definecolor{mintgreen}{RGB}{189, 255, 222} 
\definecolor{paleyellow}{RGB}{255, 255, 204} 
\definecolor{blush}{RGB}{255, 204, 229}     
\definecolor{skyblue}{RGB}{135, 206, 250}   
\definecolor{cream}{RGB}{255, 253, 208}     
\definecolor{lightcoral}{RGB}{240, 128, 128} 
\definecolor{softpurple}{RGB}{230, 204, 255} 

\definecolor{deepblue}{RGB}{0, 51, 102}        
\definecolor{richblack}{RGB}{0, 0, 0}           
\definecolor{charcoal}{RGB}{54, 69, 79}         
\definecolor{midnightblue}{RGB}{25, 25, 112}    
\definecolor{forestgreen}{RGB}{34, 139, 34}     
\definecolor{crimson}{RGB}{220, 20, 60}         
\definecolor{darkslategray}{RGB}{47, 79, 79}    
\definecolor{darkviolet}{RGB}{148, 0, 211}      
\definecolor{goldenrod}{RGB}{218, 165, 32}      
\definecolor{slateblue}{RGB}{106, 90, 205}      
\definecolor{indigo}{RGB}{75, 0, 130}           

\usepackage{adjustbox}
\usepackage{wrapfig}

\usepackage[utf8]{inputenc}
\usepackage{textcomp}

\usepackage{multirow}   

\usepackage{cleveref}

\crefname{section}{Section}{Sections}
\crefname{figure}{Figure}{Figures}
\crefname{table}{Table}{Tables}
\crefname{equation}{Eq.}{Eqs.}
\crefname{algocf}{Algorithm}{Algorithms} 
\crefname{appendix}{Appendix}{Appendices}

\Crefname{section}{Section}{Sections}
\Crefname{figure}{Figure}{Figures}
\Crefname{table}{Table}{Tables}
\Crefname{equation}{Equation}{Equations}
\Crefname{algocf}{Algorithm}{Algorithms} 
\Crefname{appendix}{Appendix}{Appendices}

\usepackage[ruled,vlined]{algorithm2e}
\SetKwProg{Fn}{Function}{}{}
\SetKwComment{tcp}{//}{}

\newcommand{\modelname}[1]{\texttt{#1}}

\title{LongWeave: A Long-Form Generation Benchmark Bridging Real-World Relevance and Verifiability}

\author{%
\textbf{Zikai Xiao}\textsuperscript{1\thanks{Co-first authors contribute equally to this work.}\thanks{Work done during an internship at Qwen Team.}}, 
\textbf{Fei Huang}\textsuperscript{2\footnotemark[1]},
\textbf{Jianhong Tu}\textsuperscript{2}, 
\textbf{Jianhui Wei}\textsuperscript{1},
\textbf{Wen Ma}\textsuperscript{1}, 
\textbf{Yuxuan Zhou}\textsuperscript{2}, \\
\textbf{Jian Wu}\textsuperscript{1},
\textbf{Bowen Yu}\textsuperscript{2},
\textbf{Zuozhu Liu}\textsuperscript{1\thanks{Corresponding authors.}},
\textbf{Junyang Lin}\textsuperscript{2\footnotemark[3]} \\ 
\\
\textsuperscript{1}Zhejiang University,
\textsuperscript{2}Qwen Team, Alibaba Group \\
\href{mailto:zuozhuliu@intl.zju.edu.cn}{\texttt{zuozhuliu@intl.zju.edu.cn}}, 
\href{mailto:junyang.ljy@alibaba-inc.com}{\texttt{junyang.ljy@alibaba-inc.com}}
}

\begin{document}
\maketitle
\begin{abstract}

Generating long, informative, and factual outputs remains a major challenge for Large Language Models (LLMs). Existing benchmarks for long-form generation typically assess real-world queries with hard-to-verify metrics or use synthetic setups that ease evaluation but overlook real-world intricacies. In this paper, we introduce \textbf{LongWeave}, which balances real-world and verifiable assessment with Constraint-Verifier Evaluation (CoV-Eval). CoV-Eval constructs tasks by first defining verifiable targets within real-world scenarios, then systematically generating corresponding queries, textual materials, and constraints based on these targets. This ensures that tasks are both realistic and objectively assessable, enabling rigorous assessment of model capabilities in meeting complex real-world constraints.
LongWeave supports customizable input/output lengths (up to 64K/8K tokens) across seven distinct tasks. Evaluation on 23 LLMs shows that even state-of-the-art models encounter significant challenges in long-form generation as real-world complexity and output length increase. Our codes are available at \href{https://github.com/ZackZikaiXiao/LongWeave}{https://github.com/ZackZikaiXiao/LongWeave}.

\end{abstract}

\section{Introduction}

Large Language Models (LLMs) have significantly enhanced their capabilities to process long inputs \citep{qwen2.5, qwen2.5-1m, grattafiori2024llama, team2023gemini} through architectural design \citep{dao2023flashattention2} and data engineering \citep{pmlr-v235-fu24d, gao2024prolong}. However, achieving robust long-sequence generation remains highly challenging \cite{que2024hellobench, bai2024longwriter}. Several research efforts have attempted to optimize LLMs for long-form output generation \citep{pham2024surimulticonstraintinstructionfollowing, bai2024longwriter, yang2024logu, xiong2025beyond}. However, the generated content often lacks adequate informativeness, comprehensiveness, and factuality~\cite{qi-etal-2024-long2rag,pradeep2024initial,song-etal-2024-veriscore}. The inherent complexity of long-form sequences further complicates accurate assessment of these qualities, highlighting the necessity for more reliable evaluation benchmarks.

\begin{figure}[t]
  \centering
  \includegraphics[width=0.42\textwidth]{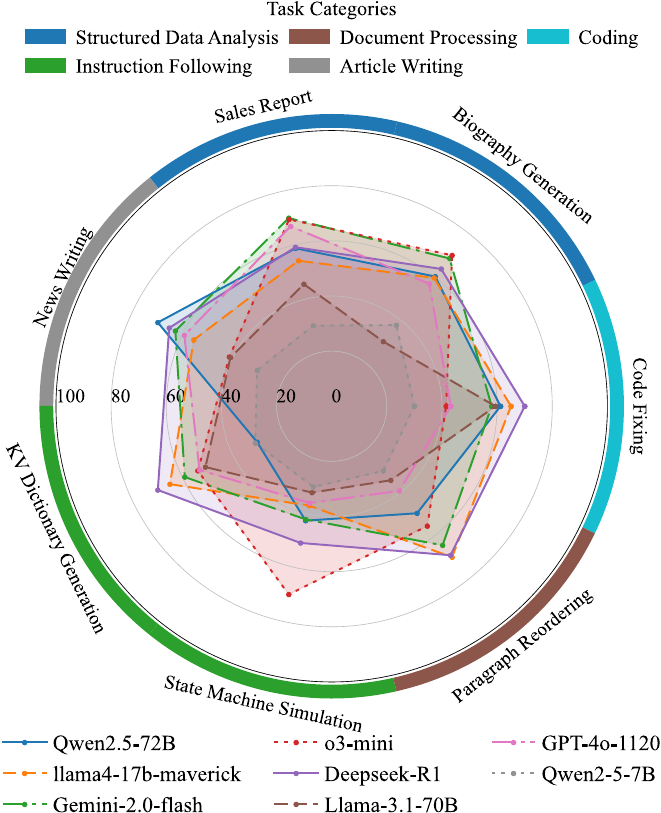}
  \caption{The performance across the seven tasks in LongWeave. For better visualization, performance scores have been normalized to a range of 0.3 to 0.7.
  }
  \label{fig:radar}
\end{figure}

Long-form generation with real-world queries is typically evaluated using similarity metrics (e.g., $\alpha$-nDCG, Self-BLEU) or LLM-as-a-Judge~\cite{bai2024longwriter}. While straightforward to implement, direct evaluation struggles with the inherent long-sequence complexity. 
To address this, another line of work breaks long-text evaluation into a set of verifiable sub-tasks, which can include factual claims (e.g., a statement like "the Earth orbits the Sun") or aspects (e.g., completeness, logical consistency). Checklists are constructed through expert-curated guidelines \cite{tan-etal-2024-proxyqa, que2024hellobench} or automated methods leveraging LLMs to extract claims from outputs for factual verification via search engines~\cite{song-etal-2024-veriscore, wei2024longform, samarinas2025beyond} or fixed databases~\cite{samarinas2025beyond}. A critical challenge lies in optimizing the degree of specificity scope: overly broad checklists produce vague claims that hinder verification, while overly detailed ones tend to over-complicate verification processes by attempting to cover all corner cases.

To enhance verifiability, some approaches use synthetic data rather than real-world data-for instance, combining short questions from datasets like MMLU~\cite{liu-etal-2024-longgenbench} into longer ones, and then checking each segment individually. Other benchmarks conduct procedural simulation or utilize objective question-answering (QA) tasks where fixed answers are associated with precise constraints to limit the response scope ~\cite{wu2025longgenbench, ye2025longproc}. Though these methods simplify verification, they generally sacrifice realism in real-world scenarios.

\begin{figure*}[htbp]
    \centering
    \includegraphics[width=1\textwidth]{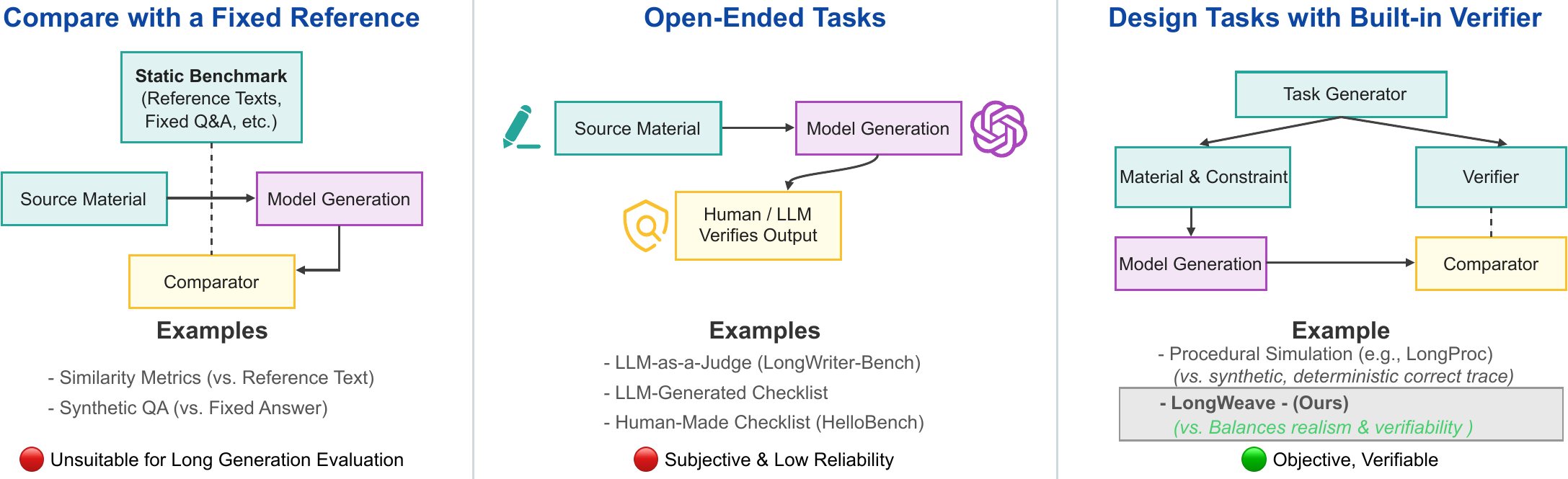}
    \caption{Three evaluation paradigms for long-form generation. LongWeave is grounded in real-world scenarios and based on objective, verifiable scoring with built-in ground truth, reducing subjectivity and inconsistencies.}
    \label{fig:pipeline}
\end{figure*}

\begin{table*}[htbp]
\centering
\footnotesize
\caption{
Comparison between long-context benchmarks. \textbf{`Open-ended'} indicates whether the task allows for diverse, creative responses. \textbf{`Deterministic'} means the task produces step-by-step, logically structured outputs. Our Constraint-Verifier Evaluation (CoV-Eval) is a constrained open-ended evaluation that synthetically constructs tasks to ensure real-world relevance. Color highlights indicate strengths (green) or challenges (orange). The length refers to the number of tokens under the cl100k tokenizer. The $\checkedstar$ symbol denotes that the characteristic is present in a subset of the benchmark's tasks.
}
\label{tab:related_benchmarks}
\setlength{\tabcolsep}{3.5pt} 
\begin{tabular}{l|ccccc}
\hline
\textbf{Benchmark} & \textbf{Input Len} & \textbf{Output Len} & \textbf{Open-ended} & \textbf{Deterministic} & \textbf{Evaluator} \\
\hline
\multicolumn{6}{c}{\textit{Benchmarks for Long Input}} \\ 
\hline
\textbf{LongBench~\cite{bai2024longbench}} & \cellcolor{softorange}$\sim$16k  & \cellcolor{softorange}$\sim$100 & \cellcolor{softgreen}$\checkedstar$ & \cellcolor{softgreen}$\checkedstar$ & Similarity\\
\textbf{RULER~\cite{hsieh2024ruler}} & \cellcolor{softgreen}$\sim$128k  & \cellcolor{softorange}$\sim$100 & \cellcolor{softorange}$\times$ & \cellcolor{softgreen}\checkmark & Rules \\
\textbf{HELMET~\cite{yen2025helmet}} & \cellcolor{softgreen}$\sim$128k  & \cellcolor{softorange}$\sim$100 & \cellcolor{softgreen}$\checkedstar$ & \cellcolor{softgreen}$\checkedstar$ & LLM-as-a-Judge\\
\textbf{InfiniteBench~\cite{zhang-etal-2024-bench}} & \cellcolor{softgreen}Infinite  & \cellcolor{softorange}$\sim$100 & \cellcolor{softorange}$\times$ & \cellcolor{softgreen}\checkmark & Rules\\
\hline
\multicolumn{6}{c}{\textit{Benchmarks for Long Generation}} \\ 
\hline
\textbf{LongWriter-Bench~\cite{bai2024longwriter}} & \cellcolor{softorange}\cellcolor{softorange}$\sim$100 & \cellcolor{softorange}$\sim$5k  & \cellcolor{softgreen}\checkmark & \cellcolor{softorange}$\times$ & LLM-as-a-Judge  \\
\textbf{LongGenBench[1]~\cite{liu-etal-2024-longgenbench}} & \cellcolor{softorange}$\sim$1k  & \cellcolor{softorange}$\sim$4k  & \cellcolor{softorange}$\times$ &\cellcolor{softgreen} \checkmark & Similarity\\
\textbf{LongGenBench[2]~\cite{wu2025longgenbench}} & \cellcolor{softorange}$\sim$100 & \cellcolor{softgreen}$\sim$8k & \cellcolor{softgreen}\checkmark &\cellcolor{softgreen} \checkmark & LLM-as-a-Judge \\
\textbf{Hello Bench~\cite{que2024hellobench}} & \cellcolor{softorange}$\sim$300  & \cellcolor{softgreen}$\sim$8k  & \cellcolor{softgreen}\checkmark & \cellcolor{softorange}$\times$ & LLM-as-a-Judge \\
\textbf{LongProc~\cite{ye2025longproc}} & \cellcolor{softgreen}$\sim$32k & \cellcolor{softgreen}$\sim$8k  & \cellcolor{softorange}$\times$& \cellcolor{softgreen} \checkmark  & Rules \\
\hline
\rowcolor{lightgray} 
\textbf{LongWeave} & \cellcolor{softgreen}\textbf{64k} & \cellcolor{softgreen}\textbf{8k} & \cellcolor{softgreen} \checkedstar & \cellcolor{softgreen}\checkedstar & \textbf{Constraint-Verifier Pairs} \\
\hline
\end{tabular}
\end{table*}

To bridge real-world relevance with verifiability, we implement decomposition at the verification stage through a new Constraint-Verifier Evaluation (CoV-Eval) mechanism, as illustrated in \cref{fig:pipeline}. Rather than extracting checklists from raw materials, which is error-prone and hard to control, CoV-Eval reverses the test construction process: it begins with predefined verifiable checklist objectives (Verifiers) grounded in real-world tasks, then synthesizes corresponding inference queries (including Constraints and materials). The Constraint acts as a constrained input that causally guides models toward generating the predefined Verifier, enabling measurable verification and evaluation. Each Constraint-Verifier (CV) pair in CoV-Eval maintains a deterministic one-to-one relationship under structurally defined rules, systematically linked to source materials. CoV-Eval contains a series of CV pairs, where each pair is linked to the corresponding material. 
These pairs can take various forms, such as a question (C) and answer (V) in QA tasks, or a triplet (C) and corresponding sentence (V) in knowledge-to-text generation, as discussed in \cref{sec:tasks}.

Based on CoV-Eval, we introduce \textbf{LongWeave}, a new benchmark evaluating five challenge scenarios of long-form generation through seven real-world relevant tasks (\cref{fig:radar}). LongWeave supports customizable input lengths (up to 64K tokens) and output lengths of 1K, 2K, 4K, and 8K tokens, with adjustable difficulty settings for each task as detailed in \cref{tab:related_benchmarks}.

Our evaluation of 23 LLMs on LongWeave reveals critical limitations in long-form generation: even top models (DeepSeek-R1) reach a performance ceiling of 54.56\%, with performance declining for 8K-token outputs (\cref{fig:radar}). Furthermore, models exhibit input-output disconnect; while supporting inputs up to 64K tokens, they fail to effectively synthesize inputs into coherent long-form responses. Expanding input context windows (e.g., to 1M tokens) does not fundamentally solve long-generation challenges and may even degrade performance. Moreover, large-scale reasoning-oriented LLMs consistently outperform general-purpose counterparts, but often suffer from failure to terminate the reasoning phase, leading to truncated outputs. Our main contributions are:

\begin{itemize}
    \item We introduce the long-form generation benchmark \textbf{LongWeave}, with CoV-Eval that bridges real-world relevance with verifiability.
    
    \item We design seven tasks, with long input sizes (up to 64K tokens), long output requirements (1-8K), and varying difficulty levels.
    
    \item Evaluation of 23 LLMs reveals critical limitations and highlights future directions in long-form generation and evaluation. 

\end{itemize}

\begin{figure*}[htbp]
    \centering
    \includegraphics[width=1.0\textwidth]{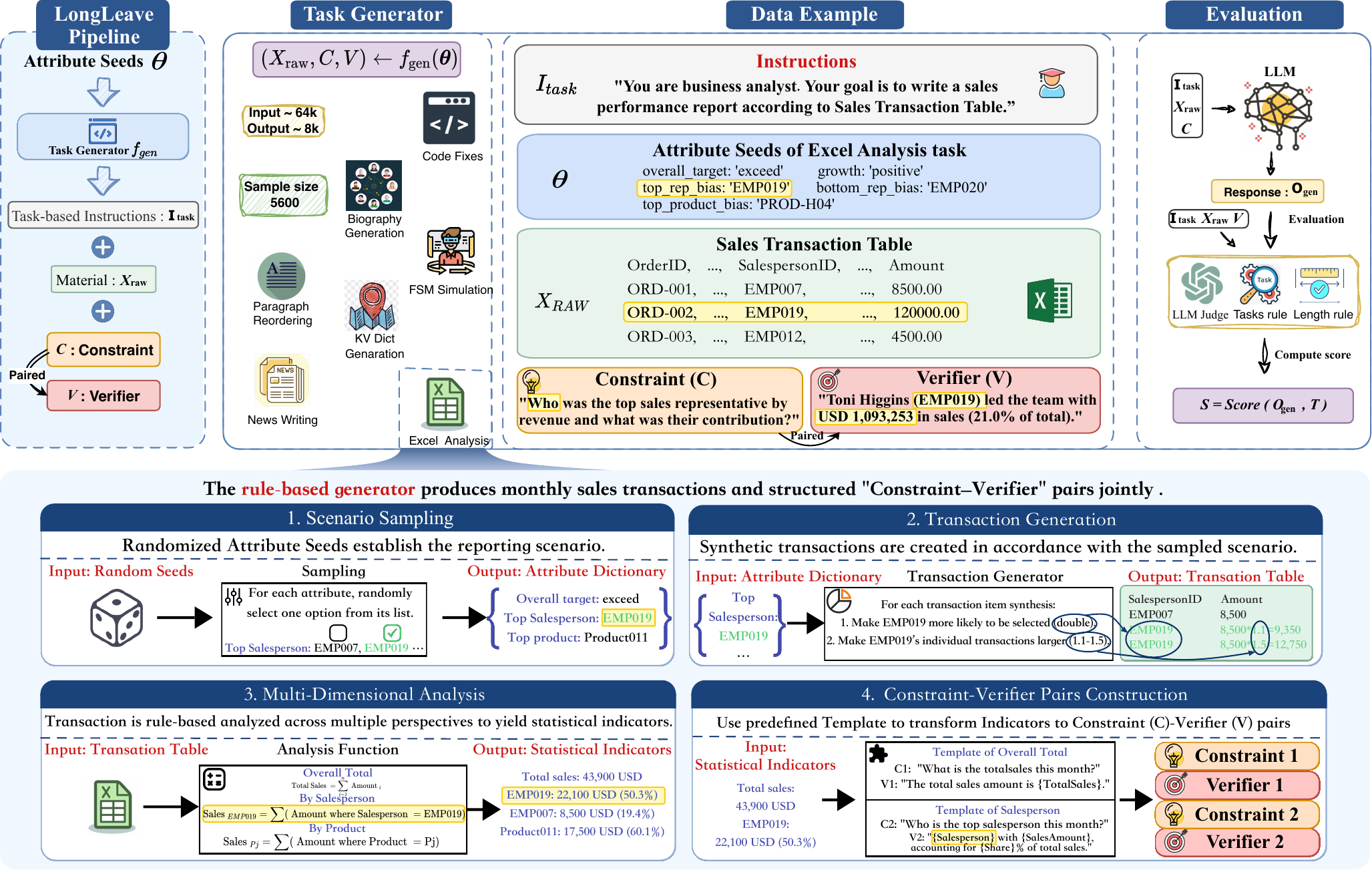}
    \caption{Illustration of the \textsc{LongLeave} evaluation pipeline. Attribute seeds define task scenarios, and the task generator creates long-form generation tasks paired with constraint–verifier sets. Model outputs are then evaluated by matching against verifiers with length and instruction-following checks.}
    \label{fig:longweave_pipeline}
\end{figure*}

\section{The LongWeave Benchmark}
In this section, we first introduce the overall pipeline of LongWeave, followed by a detailed formulation of our Constraint-Verifier Evaluation and a description of the individual tasks.

\subsection{Pipeline of LongWeave}
As shown in \cref{fig:longweave_pipeline}, the LongWeave pipeline consists of three steps: Construction, Evaluation, and Scoring. \textbf{In Construction}, task-specific attributes are systematically sampled through deterministic rule-based algorithms to generate perfectly aligned triples: (1) \textit{raw material}, (2) \textit{constraint}, and (3) \textit{verifier}. The LLM processes the material and constraints during Evaluation to produce a response that meets constraints. Finally, in the scoring phase, the output is compared to the target and then aggregated to calculate the total score.
Finally, in Scoring, the output is compared to the verifier using a scoring function and aggregated to calculate the total score.

\subsection{Constraint-Verifier-Based Evaluation (CoV-Eval)}
\noindent\textbf{Formulation of Basic Evaluation.}
We formulate long-form constrained generation as the task where an LLM, denoted as $\mathcal{L}$, must produce an output sequence $O_\text{gen}$. 
The input consists of a potentially lengthy raw material $X_\text{raw}$, and task-specific instruction $I_\text{task}$, which specifies criteria for the target output's length $\mid O_\text{gen}\mid$, content accuracy, structural formatting, and logical coherence.
The generation process is modeled as:
\begin{equation}
    O_\text{gen} = \mathcal{L}(X_\text{raw}, I_\text{task})
    \label{eq:generation1}
\end{equation}
The primary challenge lies in ensuring $O_\text{gen}$ adheres to all facets of $I_\text{task}$, especially as the input and output lengths increase, and as $I_\text{task}$ becomes more complex.




\noindent\textbf{Data Construction Stage of CoV-Eval.}
To ensure the benchmark is both realistic and verifiable, we introduce a construction process that jointly generates the raw material $X_{\text{raw}}$, the constraint $C$, and the corresponding Verifier $V$. The entire construction process is formalized as:
\begin{equation}
    (X_\text{raw}, C, V) \leftarrow f_{\text{gen}}(\boldsymbol{\theta}), \quad \text{where} \quad \boldsymbol{\theta} \sim \Theta
    \label{eq:construction}
\end{equation}
The process is driven by a \textbf{Generator} ($f_{\text{gen}}$)—a set of task-specific, deterministic, rule-based scripts, which can be seen \textbf{in the bottom part of \cref{fig:longweave_pipeline}}. The generator's behavior is controlled by structured \textbf{Attribute Seeds} ($\boldsymbol{\theta}$), which are sampled from a predefined attribute space $\Theta$ and specify properties like material scale, reasoning complexity, and constraint strictness. 
CoV-Eval combines deterministic generation with explicit attribute control, guarantees that every Constraint–Verifier pair is grounded in its material, and can be automatically verified.

\noindent\textbf{Evaluation Stage.}
The input instruction incorporates both the material and the constraint, while the output is evaluated based on whether it correctly reflects the verifier $V$ associated with the constraint $C$.
Specifically, the generation process is modeled as shown in \cref{eq:generation1}:
Specifically, under CoV-Eval, the generation process is updated as shown in \cref{eq:generation2}:
\begin{equation}
    O_\text{gen} = \mathcal{L}(X_\text{raw}, I_\text{task}, \mathcal{C}),
    \label{eq:generation2}
\end{equation}
then the quality $S$ of $O_{gen}$ is quantified by a task-specific scoring function $Score$, as shown in \cref{eq:scoring_concise}:
\begin{equation}
    S = \text{Score}(O_{gen}, V).
    \label{eq:scoring_concise}
\end{equation}
LongWeave evaluates LLMs by measuring \( S \) across diverse tasks that vary in input/output lengths and task complexity.

\subsection{Tasks}
\label{sec:tasks}
We now introduce each task, where the Constraint–Verifier pair varies by task. We indicate Material as $X_{\text{raw}}$, Constraint as $C$, and Verifier as $V$. We use rule-based scripts to generate $X_{\text{raw}}$, $C$ and $V$. For AP Style News Writing, we use LLM to generate news topics and statements. For Paragraph Reordering, original texts are from QreCC documents \cite{qrecc}.

\begin{table*}[htbp]
\centering
\caption{LongWeave Tasks: A summary of the tasks, outlining their names, abbreviations, core challenges, important configuration settings, and evaluation metrics. Metric types are color-coded as described in the table's legend. \textcolor{crimson}{Red} refers to LLM-as-a-Judge metrics. \textcolor{blue}{Blue} indicates length scores. \textcolor{darkviolet}{Purple} represents other rule-based metrics.}
\label{tab:task_list}
\small 
\setlength{\tabcolsep}{4pt} 
\renewcommand{\arraystretch}{1.2} 

\begin{tabularx}{\textwidth}{@{} 
>{\raggedright\arraybackslash}p{2.8cm} 
>{\centering\arraybackslash}p{1.5cm} 
>{\raggedright\arraybackslash}p{2.5cm} 
>{\raggedright\arraybackslash}X 
>{\raggedright\arraybackslash}X 
@{}}
\toprule
\multicolumn{1}{@{}l}{\textbf{Task Name}} & 
\multicolumn{1}{c}{\textbf{Abbrev.}} & 
\multicolumn{1}{l}{\textbf{Challenge}} & 
\multicolumn{1}{l}{\textbf{Configuration}} & 
\multicolumn{1}{l@{}}{\textbf{Metrics}} \\
\midrule

Code Fixing with Flake8 Compliance & CF & Coding & 
violation\_prob = 0.85 \newline
\textbf{error\_lines $\propto$ gen\_len}
& \textcolor{darkviolet}{Runnability} \newline 
\textcolor{crimson}{Style score} \newline
\textcolor{blue}{length score} \\

\hdashline

KG to Text Biography Generation & BioG & Structured Data Analysis & 
\textbf{triple\_count $\propto$ gen\_len}
& \textcolor{crimson}{Coverage Rate}. \\
\hdashline

CSV Sales Report Analysis & SR & Structured Data Analysis & 
\textbf{record\_count $\propto$ gen\_len} \newline
\textbf{target\_count $\propto$ gen\_len}

& \textcolor{crimson}{Coverage Rate} \newline
\textcolor{crimson}{Correctness Rate} \\
\hdashline

AP Style News Writing & NW & Article Writing & 
\textbf{fact\_counts $\propto$ gen\_len} \newline
ap\_stylebook\_rules  \newline
& \textcolor{crimson}{Coverage Rate} \newline
\textcolor{crimson}{Style Score} \\
\hdashline

KV Dictionary \newline Generation & KVG & Instruction Following & 
\textbf{entry\_count $\propto$ gen\_len} \newline
key\_length = 32 \newline
value\_length = 32
& \textcolor{darkviolet}{Existence Score} \newline
\textcolor{blue}{Length score} \newline
\textcolor{darkviolet}{Position score} \\
\hdashline

State Machine Simulation & SMS & Instruction Following & 
num\_states = 3 \newline
input\_size = 3 \newline
output\_size = 3 \newline
\textbf{step\_length $\propto$ gen\_len }
& \textcolor{darkviolet}{Step Match Ratio} \\
\hdashline

Paragraph Reordering & PR & Document Processing & 
\textbf{para\_length $\propto$ gen\_len }
& \textcolor{darkviolet}{Kendall's Tau}. \\
\bottomrule
\end{tabularx}

\end{table*}

\noindent\textbf{Code Fixing.} This task requires LLMs to \textit{fix Python code with Flake8 style} violations (line length, indentation) while ensuring the code remains runnable. We design the \textit{code polluter} to inject Flake8 violations into a randomly generated runnable Python code, forming a polluted code ($X_{\text{raw}}$). The LLM is prompted to fix the code. The part code required repair is $C$. The repaired code can be automatically checked by Flake8 toolkit($V$).

\noindent\textbf{KG to Text Biography Generation (BioG).} This task evaluates LLMs' ability to generate coherent and factual biographies based on given knowledge graph (KG) triples. The designed \textit{knowledge graph generator} creates a large set of task relationships around a central person, then extracts triples (subject-predicate-object) and corresponding sentences starting from the nearest nodes. 
The evaluated model needs to incorporate all triples ($C$) into a fluent narrative within the specified word count. The verifier ($V$) is a rule-based natural language statement derived from these triples. The model is evaluated on its ability to accurately integrate all triples into the generated text, with penalties for missing or fabricated information.

\noindent\textbf{CSV Sales Report Analysis (SR).} This task evaluates LLMs' ability to generate a sales report and answer predefined, specific questions based on a transaction table. We designed a \textit{sales report generator} that synthesises the transaction table ($X_{\text{raw}}$), while generating natural language questions ($C$) and corresponding answers ($V$). Evaluation focuses on both coverage and accuracy of the answers.

\noindent\textbf{AP Style News Writing (NW).} 
This task evaluates LLMs' ability to write a news article following the Associated Press Stylebook (AP Style)~\cite{goldstein1998associated}. Given a news topic ($X_{\text{raw}}$), GPT-4o-2024-11-20 generates correct fact statements ($V$) together with corresponding flawed statements ($C$) that violate AP Style rules.
The evaluated LLM is then required to write an article on the topic, integrating all statements in their correct form.

\noindent\textbf{KV Dictionary Generation (KVG).}
This task, the inverse of KV Retrieval in~\cite{hsieh2024ruler}, evaluates LLMs' ability to generate a dictionary string with a target key–value pair placed \textit{at the correct index}, following formatting rules (e.g., keys in uppercase with underscores); the query specifying the key–value pair and index is the \textit{Constraint} ($C$), and a rule-based script verifies placement and formatting as the \textit{Verifier} ($V$).

\noindent\textbf{State Machine Simulation (SMS).} 
This task requires simulating state transitions of a finite state machine (FSM)~\cite{lee1996principles} step by step. Here, the transition table serves as the raw material ($X_{\text{raw}}$), the initial state and input string constitute the \textit{Constraint} ($C$), and an FSM validation script acts as the \textit{Verifier} ($V$) by checking the generated sequence against the correct state transitions and signals. Models are evaluated by their match ratio and overall accuracy in reproducing all steps without errors.

\noindent\textbf{Paragraph Reordering (PR).} This task requires LLMs to reorder shuffled paragraphs ($C$) into the coherent sequence ($V$). The material consists of randomly sampled paragraphs, with the constraints being the shuffled order and the verifier being the correct sequence. Evaluation uses Kendall's Tau to measure the consistency of the predicted order~\cite{liu-etal-2020-evaluating, shen-baldwin-2021-simple}.



\subsection{Input Length Statistic}
\begin{figure}[htbp]
  \centering
  \includegraphics[width=0.45\textwidth]{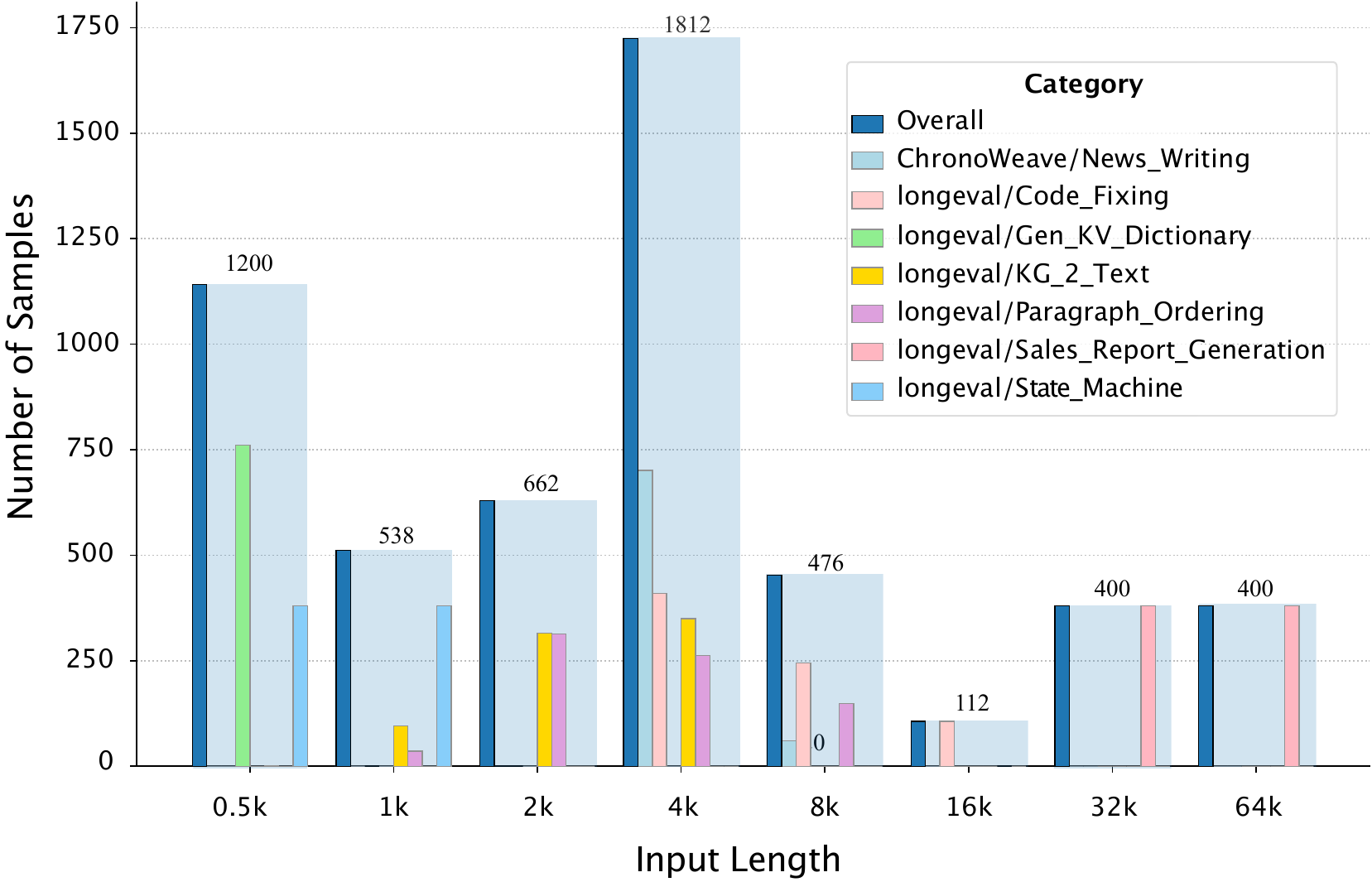}
  \caption{Input length distribution of LongWeave}
  \label{fig:input_len_dist}
\end{figure}

Generative tasks with long input contexts are critical yet underexplored. To reduce hallucinations, users often provide extensive context for generating complex outputs. Unlike prior benchmarks capped at ~1k tokens\citep{bai2024longwriter, wu2025longgenbench, liu-etal-2024-longgenbench, que2024hellobench}, LongWeave supports up to 64k-token inputs, enabling evaluation in real-world scenarios like structured file analysis and document processing. We provide the input length distribution of LongWeave in \cref{fig:input_len_dist}. 

\subsection{Output Length Control}
LongWeave controls target output length through a multi-faceted strategy. First, the scale and complexity of input materials—such as the number of data records or code lines—are procedurally generated to be proportional to each target length tier (1K, 2K, 4K, 8K tokens). Second, prompts provide models with explicit instructions specifying the desired output length. Finally, the evaluation protocol enforces these constraints by directly penalizing length deviations.

\subsection{Evaluation Metrics}
The evaluation metrics include LLM-as-a-Judge for CoV-Eval, length-related metrics, and others. Each task’s final score is the harmonic mean of sub-metrics, ensuring that poor performance in one area cannot be offset by better performance in another.

\noindent \textbf{LLM-as-a-Judge} use LLM-as-a-Judge to check whether verifiers, corresponding to defined constraints, are accurately reflected in the model's output. These include style, factual coverage, and question answering. The \textit{Style Score} measures adherence to Flake8 standards, penalizing unresolved violations. The \textit{Factual Coverage Rate} tracks the proportion of knowledge graph triples in the text, while the \textit{Answer Coverage Rate} measures the proportion of answered analytical questions. The \textit{Correctness Rate} calculates answer accuracy, and the \textit{Factual Statement Coverage Rate} tracks recall of required factual statements. The \textit{AP Style Score} quantifies adherence to AP Stylebook guidelines.

\noindent \textbf{Length Score} is used to test whether the model outputs according to the required length. The implicit length score is applied when truncation occurs after exceeding the length, while the explicit length score is used in CF and KVG, where the length score is treated as a sub-score.

\noindent \textbf{Rule-based metrics} rely on deterministic code to verify correctness, such as code runnability, existence and placement of target key–value pairs, and Kendall’s Tau for paragraph reordering.

\section{Experiments}

\subsection{Models and Inference Setup}
\label{sec:models_inference_setup_stable}

We evaluated a range of LLMs using LongWeave, comprising proprietary and commercial API-accessed models, open-source models, and reasoning models. The \textit{long-generation models} assessed include LongWriter-glm4-9B \cite{glm2024chatglm, bai2024longwriter}. The \textit{open-source models} include the Llama-3-series, Llama-4-series \cite{grattafiori2024llama}, Phi-4-mini-instruct, Qwen2.5-series (3B, 7B, 14B, 72B, QwQ-Plus) \cite{qwen2.5}, and the newer Qwen3 series (4B, 8B, 14B-Think/Non-Think, 32B-Think/Non-Think) \citep{qwen3}. Additionally, we evaluated DeepSeek-V3 \cite{liu2024deepseek}. The \textit{commercial models} include GPT-4o-2024-11-20 \cite{achiam2023gpt}, Gemini-2.0-flash \cite{team2023gemini}, and Qwen-long. Specialized \textit{reasoning models}, such as o3-mini-2025-01-31 and DeepSeek-R1 \cite{guo2025deepseek}, were also included in the evaluation. The open-source model uses VLLM deployment on A100.

\subsection{Task Configurations}
\label{sec:task_configurations_ultra_concise}
LongWeave evaluates LLMs across seven distinct tasks, each with four variants targeting output lengths of 1k, 2k, 4k, and 8k tokens. For each variant, 200 test samples are used, resulting in a total of 5,600 samples per model. We primarily used \modelname{Qwen2.5-72B-Instruct} for LLM-as-a-Judge evaluations. To control output length, we adjust the configuration as illustrated by the "gen\_len" configurations in \cref{tab:task_list}. Furthermore, LongWeave supports task difficulty control through adjustments to input complexity (e.g., key\_length in KVG), the strictness of constraints (e.g., AP stylebook rules in NW), and structural requirements of the target output (e.g., step\_length, para\_length).

\begin{figure}[t]
  \centering
  \includegraphics[width=0.45\textwidth]{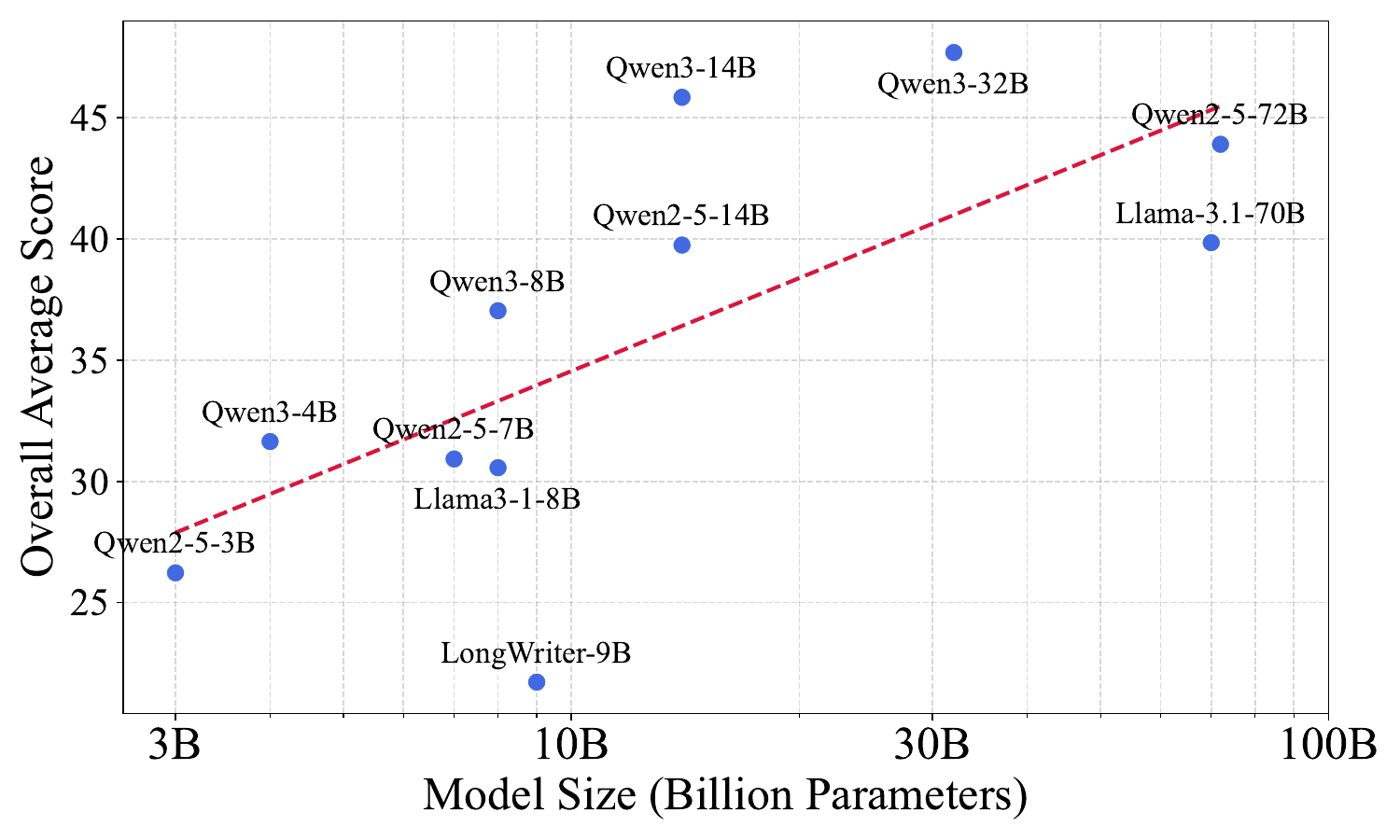}
  \caption{Performance of different model sizes}
  \label{fig:scaling_law}
\end{figure}

\begin{table*}[htbp]
  \centering
    \caption{Model performance summary (task-average and length-average scores). The highest model performance for each task and score is bolded, and for the overall performance, models in the top 5 are bolded.}
    
  \label{tab:model_summary}
  
  \scriptsize  

  \resizebox{\textwidth}{!}{
  \begin{tabular}{l*{12}{S}}
    \toprule
    & \multicolumn{7}{c}{\textbf{Task scores}} &
      \multicolumn{4}{c}{\textbf{Length scores}} &
      \textbf{Overall} \\
    \cmidrule(lr){2-8}\cmidrule(lr){9-12}
    \textbf{Model} &
      {CF} & {BioG} & {SR} & {NW} & {KVG} & {SMS} & {PR} &
      {1k} & {2k} & {4k} & {8k} &
      {Avg} \\
    \midrule

    LongWriter-glm4-9B                          & 29.67 & 67.27 & 18.23 & 14.62 & 04.48 & 03.68 & 13.99 & 24.55 & 23.11 & 20.69 & 18.48 & 21.71 \\
    Phi-4-mini-Instruct                         & 00.02 & 69.86 & 10.50 & 18.30 & 03.62 & 03.25 & 39.51 & 23.64 & 20.27 & 20.58 & 18.40 & 20.72 \\
    Llama3-1-8B-Instruct                        & 46.76 & 60.66 & 13.93 & 20.29 & 15.97 & 3.82 & 52.46 &40.11&	34.75&	27.13&	20.25&	30.56 \\
    Llama3-1-70B-Instruct                 & 58.45 & 69.36 & 20.04 & 24.08 & 40.35 & 06.43 & 60.26 & 53.58 & 46.92 & 33.85 & 25.07 & 39.85 \\
    Llama4-scout-17B-16e-Instruct              & 33.24 & 76.38 & 24.47 & 28.25 & 33.32 & 06.20 & 67.88 & 48.65 & 37.55 & 37.22 & 30.72 & 38.53 \\
    Llama4-17b-128e-Instruct          & 64.63 & 84.09 & 22.94 & 27.96 & 55.11 & 09.84 & \textbf{91.51} & 55.81 & 54.93 & 50.61 & 42.12 & \textbf{50.87} \\

    Qwen2.5-3B-Instruct                         & 16.33 & 66.42 & 09.82 & 20.27 & 20.82 & 02.85 & 47.05 & 30.64 & 28.16 & 24.76 & 21.33 & 26.22 \\
    Qwen2.5-7B-Instruct                         & 26.09 & 73.16 & 14.91 & 21.27 & 19.64 & 04.94 & 56.45 & 38.13 & 33.77 & 27.19 & 24.60 & 30.92 \\
    Qwen2.5-14B-Instruct                        & 49.48 & 80.94 & 19.33 & 23.80 & 22.80 & 05.72 & 76.18 & 47.60 & 42.97 & 36.07 & 32.35 & 39.75 \\
    Qwen2.5-72B-Instruct                        & 60.43 & 84.41 & 24.48 & \textbf{31.76} & 18.84 & 13.77 & 73.67 & 51.67 & 48.69 & 40.99 & 34.29 & 43.91 \\

    Qwen3-4B                                    & 28.36 & 73.29 & 17.89 & 18.19 & 24.87 & 11.87 & 47.02 & 44.28 & 35.92 & 27.18 & 19.18 & 31.64 \\
    Qwen3-8B                                    & 45.92 & 76.88 & 18.98 & 18.90 & 17.70 & 13.40 & 67.50 & 46.86 & 40.08 & 34.06 & 27.17 & 37.04 \\
    Qwen3-14B                                   & 59.10 & 79.00 & 21.96 & 22.12 & 33.88 & 18.45 & 86.43 & 56.87 & 49.34 & 42.28 & 34.91 & 45.85 \\
    Qwen3-32B                                   & 63.44 & 79.77 & 24.95 & 21.46 & 44.36 & 16.18 & 83.71 & 59.71 & 52.68 & 44.57 & 33.82 & 47.70 \\

    DeepSeek-V3                                 & 59.43 & 80.62 & 23.25 & 27.11 & 33.25 & 11.47 & 91.12 & 56.30 & 51.61 & 43.19 & 35.34 & 46.61 \\
    Qwen-long                                   & 35.78 & 77.65 & 24.87 & 26.67 & 27.78 & 12.68 & 78.88 & 49.50 & 44.03 & 39.15 & 29.78 & 40.62 \\
    GPT-4o-2024-11-20                           & 40.60 & 82.72 & 27.20 & 28.96 & 42.82 & 09.16 & 64.58 & 56.18 & 50.30 & 37.65 & 25.03 & 42.29 \\
    Gemini-2.0-flash                            & 56.93 & 88.58 & 28.22 & 29.91 & 48.94 & 13.46 & 86.68 & 60.44 & 56.17 & 49.20 & 35.75 & \textbf{50.39} \\

    \hdashline
    DeepSeek-R1-Distill-Qwen-7B                 & 00.00 & 47.19 & 04.77 & 11.76 & 05.62 & 02.39 & 30.50 & 18.63 & 13.43 & 14.21 & 12.14 & 14.60 \\
    DeepSeek-R1-Distill-Qwen-32B                & 54.14 & 66.65 & 22.25 & 21.31 & 14.54 & 08.86 & 73.70 & 45.06 & 39.59 & 35.74 & 29.01 & 37.35 \\
    Qwen3-14B-Think                                   & 45.41 & 78.71 & 28.59 & 23.86 & 42.92 & 10.64 & 89.30 & 52.33 & 49.96 & 43.96 & 36.30 & 45.64 \\
    Qwen3-32B-Think                                   & 59.69 & 83.64 & \textbf{35.67} & 20.86 & 52.40 & 13.89 & 88.38 & 57.71 & 56.89 & 49.54 & 38.45 & 50.65 \\
    DeepSeek-R1                                 & \textbf{70.10} & 86.16 & 24.62 & 30.56 & \textbf{60.14} & 19.60 & 90.73 & \textbf{63.86} & \textbf{59.25} & \textbf{52.85} & \textbf{42.28} & \textbf{54.56} \\
    QwQ-plus-2025-03-05                         & 57.22 & 80.71 & 26.66 & 25.66 & 40.96 & 26.10 & 85.04 & 62.40 & 51.82 & 44.20 & 37.21 & \textbf{48.91} \\
    o3-mini-2025-01-31                          & 38.76 & \textbf{89.30} & 28.06 & 24.21 & 43.51 & \textbf{33.06} & 78.88 & 62.06 & 56.12 & 43.04 & 30.66 & \textbf{47.97} \\
    
    \bottomrule
  \end{tabular}
  }
\end{table*}

\begin{table}[htbp]
\centering
\small
\caption{Distribution of failure patterns across 1,400 analyzed samples, grouped by category.}
\label{tab:failure_patterns}
\begin{tabular}{l r r}
\toprule
\textbf{Failure Pattern} & \textbf{Count} & \textbf{Share} \\
\midrule
\multicolumn{3}{l}{\textit{Instruction-following errors}} \\
\quad Selective instruction execution & 375 & 30.4\% \\
\quad Stepwise deviation & 172 & 14.0\% \\
\quad Incomplete factual coverage & 156 & 12.7\% \\
\quad Length control issues & 153 & 12.4\% \\
\midrule
\multicolumn{3}{l}{\textit{Numerical errors}} \\
\quad Calculation failures & 123 & 10.0\% \\
\midrule
\multicolumn{3}{l}{\textit{Content problems}} \\
\quad Fabricated facts & 27 & 2.2\% \\
\quad Redundancy / looping & 17 & 1.4\% \\
\midrule
\multicolumn{3}{l}{\textit{Reasoning-specific failures}} \\
\quad Failure to terminate reasoning & 210 & 17.0\% \\
\bottomrule
\end{tabular}
\end{table}

\subsection{Main Results}
The results are summarized in ~\cref{tab:model_summary}. In the table, we have divided all the models into standard models and reasoning models. We have listed the average performance across seven tasks at four different input lengths, as well as the overall average performance across all tasks at four lengths.

\noindent \textbf{Existing models struggle in long form generation.}
Frontier proprietary models demonstrate the best performance. DeepSeek-R1, Gemini-2.0-flash, and o3-mini-2025-01-31 achieve nearly 60\% performance at 1k length, but when generating 8k, the performance drops to around 40\%. GPT-4o-2024-11-20 only achieves 42.99\% while it tends to generate short responses.

\noindent \textbf{Increasing model size can improve long generation quality.}  
Llama4-17b-128e achieves the best performance due to having the largest number of parameters. The three smallest models, Phi-4-mini, Qwen-2.5-3B, and Qwen3-4b, all perform below 30\%. We visualize the relationship between model size and corresponding performance in \cref{fig:scaling_law}, where the regression curve shows a positive correlation between the two.

\noindent \textbf{Performance of Reasoning Models on Long-Sequence Generation.} The very large reasoning models (e.g., \textsc{DeepSeek-R1}) perform strongly on long-sequence generation tasks (\cref{tab:model_summary}). In contrast, smaller reasoning models often fail to terminate the reasoning phase, repeatedly generating large chunks of the input, which leads to truncation.

\noindent \textbf{Performance Degradation when Input Context is Long.} As shown in \cref{tab:model_summary}, the quality of long outputs deteriorates significantly with longer inputs. This is especially evident in tasks like Sales Report Analysis and writing AP-style News Articles, which require handling long materials and detailed guidelines. Despite this challenge, managing both long inputs and outputs is crucial for practical applications. By incorporating more relevant information into the input window, hallucinations can be minimized, offering a key direction for optimizing long-sequence generation models.

\begin{table*}[htbp]
\centering
\caption{Performance comparison across different tasks under varying sample sizes. Values represent mean performance metrics with standard deviations (format: $mean_{\pm std}$).}
\label{tab:stable_test_performance}
\scalebox{0.85}{
\footnotesize
\setlength{\tabcolsep}{3.5pt}
\begin{tabular}{@{}l*{9}{c}c@{}}
\toprule
\textbf{Task} & \multicolumn{10}{c}{\textbf{Number of Samples}} \\
\cmidrule(lr){2-11}
& 20 & 40 & 60 & 80 & 100 & 120 & 140 & 160 & 180 & 200 \\
\midrule
CF      & $36.23_{\pm 2.72}$ & $35.31_{\pm 2.80}$ & $35.88_{\pm 2.50}$ & $36.12_{\pm 2.30}$ & $37.41_{\pm 0.65}$ & $36.65_{\pm 0.55}$ & $36.45_{\pm 0.45}$ & $36.37_{\pm 0.50}$ & $36.82_{\pm 0.32}$ & $36.97_{\pm 0.28}$ \\
BioG    & $60.42_{\pm 0.16}$ & $61.21_{\pm 0.18}$ & $61.10_{\pm 0.20}$ & $61.15_{\pm 0.18}$ & $61.03_{\pm 0.28}$ & $61.08_{\pm 0.25}$ & $61.20_{\pm 0.23}$ & $61.48_{\pm 0.30}$ & $61.35_{\pm 0.28}$ & $61.14_{\pm 0.41}$ \\
SR      & $13.16_{\pm 0.25}$ & $12.26_{\pm 0.30}$ & $12.60_{\pm 0.28}$ & $12.80_{\pm 0.24}$ & $13.81_{\pm 0.29}$ & $13.50_{\pm 0.22}$ & $13.60_{\pm 0.18}$ & $13.86_{\pm 0.20}$ & $14.05_{\pm 0.16}$ & $14.13_{\pm 0.14}$ \\
NW      & $20.03_{\pm 0.01}$ & $19.06_{\pm 0.10}$ & $19.40_{\pm 0.15}$ & $19.50_{\pm 0.18}$ & $19.75_{\pm 0.20}$ & $19.85_{\pm 0.18}$ & $19.90_{\pm 0.16}$ & $19.37_{\pm 0.30}$ & $19.55_{\pm 0.35}$ & $19.62_{\pm 0.48}$ \\
KVG     & $14.55_{\pm 1.47}$ & $13.80_{\pm 1.50}$ & $14.00_{\pm 1.40}$ & $14.30_{\pm 1.20}$ & $15.29_{\pm 0.17}$ & $14.60_{\pm 0.30}$ & $14.80_{\pm 0.25}$ & $14.20_{\pm 0.30}$ & $14.95_{\pm 0.45}$ & $15.20_{\pm 0.68}$ \\
SMS     & $3.19_{\pm 0.07}$  & $3.66_{\pm 0.10}$  & $3.60_{\pm 0.09}$  & $3.63_{\pm 0.08}$  & $3.69_{\pm 0.03}$  & $3.70_{\pm 0.02}$  & $3.74_{\pm 0.02}$  & $3.74_{\pm 0.02}$  & $3.78_{\pm 0.02}$  & $3.81_{\pm 0.01}$  \\
PR      & $54.67_{\pm 0.13}$ & $58.02_{\pm 0.20}$ & $58.10_{\pm 0.15}$ & $58.15_{\pm 0.12}$ & $57.23_{\pm 0.86}$ & $57.90_{\pm 0.80}$ & $58.10_{\pm 0.75}$ & $58.02_{\pm 0.10}$ & $59.20_{\pm 0.15}$ & $60.05_{\pm 0.12}$ \\
\midrule
Overall & $28.92_{\pm 0.30}$ & $29.04_{\pm 0.35}$ & $29.10_{\pm 0.40}$ & $29.30_{\pm 0.35}$ & $29.80_{\pm 0.01}$ & $29.60_{\pm 0.10}$ & $29.80_{\pm 0.15}$ & $29.69_{\pm 0.10}$ & $30.05_{\pm 0.12}$ & $30.13_{\pm 0.11}$ \\
\bottomrule
\end{tabular}
}
\end{table*}

\subsection{Failure Pattern}
To better understand where models fail, we analyzed 1,400 samples across seven tasks and four target lengths (1k, 2k, 4k, and 8k; 200 samples per task and 50 per length). Outputs were generated with \modelname{Qwen3-32B} in think mode (without thinking budgets). Failure types were first labeled with \modelname{GPT-5-2025-08-07} and then manually checked. We observed eight common failure patterns, which we group into four categories (Table~\ref{tab:failure_patterns}).

\noindent \textbf{Instruction-following errors} are the most common. Selective instruction execution (30.4\%) means models handle the easy parts of a prompt but ignore the harder constraints. Stepwise deviations (14.0\%) show that the ability to follow instructions gets worse as the output becomes longer. Incomplete factual coverage (12.7\%) often happens in structured tasks like \textsc{BioG}, where some required facts are dropped. Length control issues (12.4\%) also appear, with outputs not matching the requested length.

\noindent \textbf{Numerical errors} (10.0\%) mostly occur in quantitative tasks like \textsc{SR}, where models miscalculate percentages, averages, or growth rates. These errors suggest that reliable number handling may require tool support.

\noindent \textbf{Content problems} include fabricated facts (2.2\%), where models add unsupported information, and redundancy or looping (1.4\%), where outputs repeat content or drift into filler text.

\noindent \textbf{Reasoning-specific failures} are unique to reasoning models. In 17.0\% of cases, the reasoning phase did not stop: models produced very long “thinking” traces, often repeating large chunks of the input, leaving too little budget for the final answer and causing truncation.

\section{Analysis}
\subsection{Stability of the Benchmark}
To assess the stability of the benchmark, we conducted multiple experiments using the Llama-3.1-8B model with varying sample sizes (20-200), as shown in \cref{tab:stable_test_performance}. We found that as the sample size increased, the total score gradually stabilized, and the variance decreased from 0.3 to 0.11. Once the sample size exceeded 100, the results converged within a margin of 0.15. For the official evaluation, we used 200 samples to ensure the stability of the benchmark's total score.


\begin{table}[tbp]
\centering
\caption{Evaluation of LLM-as-Judge Stability in CoV-Eval using Different Scoring Models}
\label{tab:llm_judge_stability}
\scalebox{0.6}{
\begin{tabular}{@{}l*{6}{c}@{}}
\toprule
 & \multicolumn{6}{c}{Scoring Models} \\
\cmidrule(lr){2-7}
Tasks & \rotatebox{0}{\shortstack{DeepSeek-\\V3}} & \rotatebox{0}{\shortstack{o3-\\Mini}} & \rotatebox{0}{\shortstack{4o-\\1120}} & \rotatebox{0}{\shortstack{Qwen-2-5\\72B}} & \rotatebox{0}{\shortstack{Qwen 2.5\\32B}} & \rotatebox{0}{\shortstack{Qwen-2-5\\14B}} \\
\midrule
CF & 51.64 & 45.2 & 40.96 & 46.76 & 3.53 & 0.76 \\
BioG & 59.86 & 59.87 & 60.13 & 60.66 & 58.01 & 57.88 \\
SR & 13.52 & 13.6 & 11.7 & 13.93 & 13.7 & 21.71 \\
NW & 19.95 & 10.4 & 28.88 & 20.29 & 10.39 & 10.67 \\
\midrule
Total Score & 31.75 & 30.14 & 31.27 & 30.56 & 23.27 & 24.04 \\
\bottomrule
\end{tabular}
}
\end{table}
\subsection{Stability of LLM-as-a-Judge}
For the CF, BioG, SR, and NW tasks, we used the Qwen-2.5-72B model as an LLM judge. To test the reliability, we used other LLMs as evaluators. The Llama-3.1-8B model was tested as the evaluated model on 100 samples with different evaluation models. As shown in \cref{tab:llm_judge_stability}, the results revealed a performance fluctuation variance of 0.45 for models like DeepSeek-V3, GPT-4o-2024-11-20, and o3-mini-2025-01-31.

\begin{figure*}[htbp]
    \centering
    \includegraphics[width=1.0\linewidth, keepaspectratio]{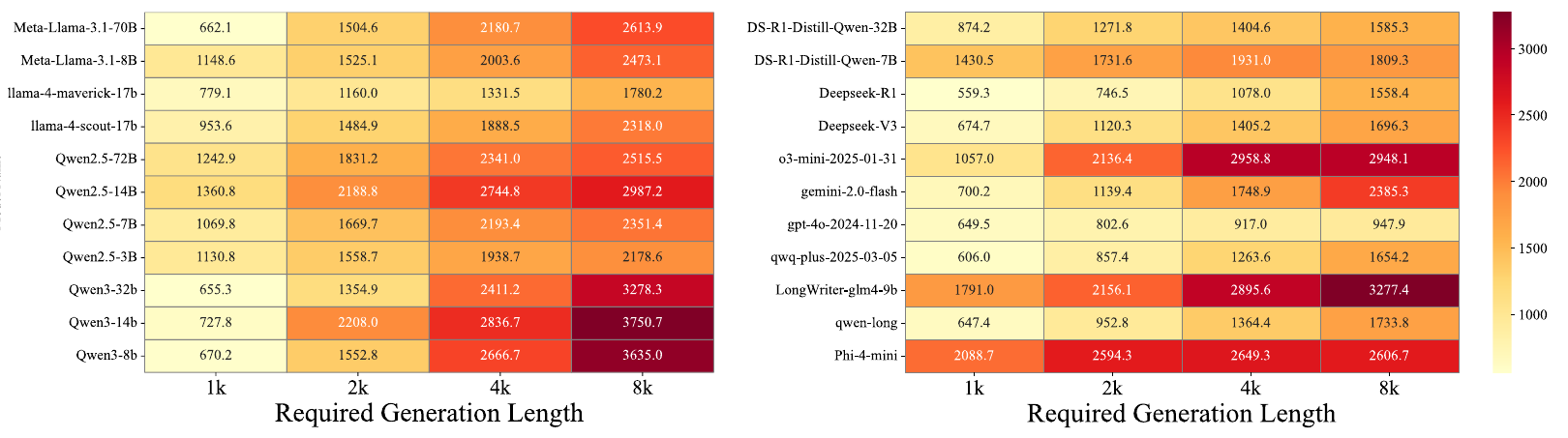}
    \caption{The heatmap visualizes the actual output lengths across four target length settings (1k, 2k, 4k, and 8k tokens). Each row represents the model, while the column corresponds to the length.}
    \label{fig:output_len_dist}
\end{figure*}

\subsection{Output Length Distribution}
During inference, we provided the models with required word counts and analyzed the output word lengths, categorizing them into four ranges: below 1k, 1k-2k, 2k-4k, and 4k-8k, as shown in \cref{fig:output_len_dist}. It was observed that, with the exception of the 03-mini, other \textit{reasoning models tend to generate shorter outputs after processing}. In contrast, smaller open-source models tend to generate longer outputs, despite their overall performance scores not being as high, indicating that output quality is not directly correlated with length. Notably, the Qwen-3 series demonstrates better length-following ability compared to the Qwen-2.5.

\subsection{Increasing the Context Window Does Not Necessarily Improve Long Generation}
\begin{figure}[htbp]
  \centering
  \includegraphics[width=0.48\textwidth]{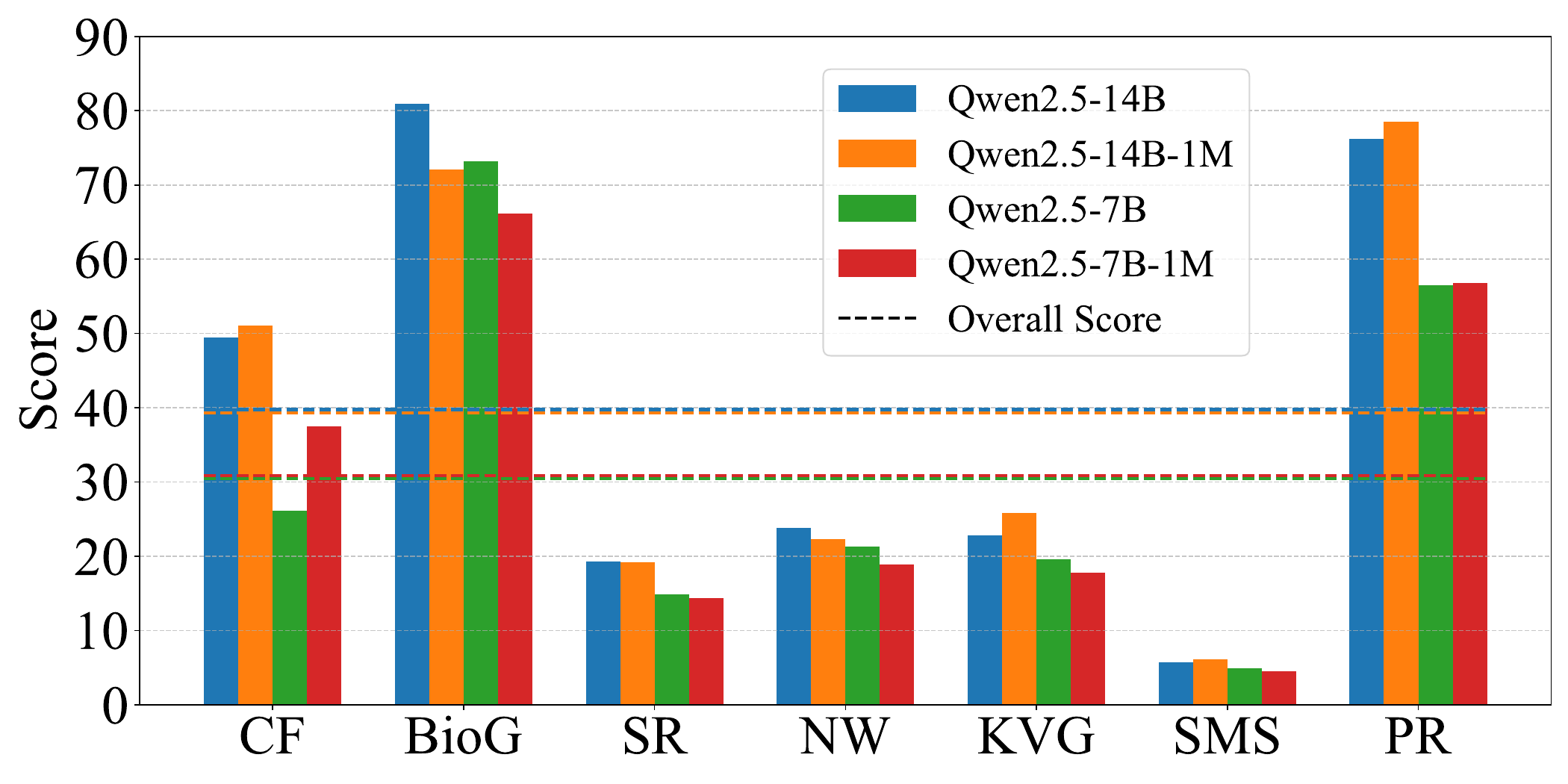}
  \caption{Input length distribution of LongWeave}
  \label{fig:1m_vs_standard}
\end{figure}

We compared the performance of the Qwen2.5-14B and 7B models with a 1M context window version, as shown in \cref{fig:1m_vs_standard}, \textit{there was little difference in overall scores}: long-input models performed better than standard models at 1K, 2K, and 4K lengths but showed decreased performance at 8K when generating ultra-long sequences. It indicates that although long input models and long generation models share the same model structure, the performance is inconsistent due to the training data.

\section{Conclusion}
Evaluating long, constrained LLM outputs is challenging. We introduce LongWeave, featuring CoV-Eval to bridge real-world relevance with objective verifiability. This suite spans seven tasks across five domains with customizable input/output lengths. Our evaluation of 23 LLMs using LongWeave demonstrates that even top models falter for long generations, with performance degrading significantly as length rises; reasoning models, however, navigate these challenges more effectively. LongWeave thereby provides a precise instrument to diagnose these systemic issues and guide the development of truly capable long-form generation.

\section*{Limitations}
While LongWeave and CoV-Eval contain several limitations that should be acknowledged:
\noindent \textbf{High Cost for Inference.} The nature of LongWeave, involving long input materials (up to 64K tokens) and the generation of long outputs (up to 8K tokens), inherently makes evaluating a wide range of models computationally expensive.
\noindent \textbf{High Cost LLM-as-a-Judge.} Several tasks within LongWeave rely on large LLMs (e.g., Qwen2.5-72B-Instruct) as judges, which adds significant computational overhead and cost.
\noindent \textbf{Limited Coverage of Creative Tasks.} LongWeave currently focuses on factual accuracy and structural correctness, but it could be expanded to better assess creative tasks.

\section*{Acknowledgments}
This work is supported by the National Key R\&D Program of China (Grant No. 2024YFC3308304), the "Pioneer" and "Leading Goose" R\&D Program of Zhejiang (Grant No. 2025C01128), 
the National Natural Science Foundation of China (Grant No. 62476241), the Natural Science Foundation of Zhejiang Province, China (Grant No. LZ23F020008), the Zhejiang Zhejiang Key Laboratory of Medical Imaging Artificial Intelligence.

\bibliography{custom}

\appendix

\section{Appendix}
\label{sec:appendix}

\subsection{Related Work}

\noindent \textbf{Long Input and Output Models}
Recent advancements in LLMs have significantly improved long-context input processing through techniques such as efficient attention (e.g., Flash Attention \citep{dao2023flashattention2}, Ring Attention \citep{liu2024ringattention}), sparse attention methods (e.g., shifted sparse attention in LongLoRA \citep{longlora}, dilated attention \citep{ding2023longnet}), and memory mechanisms like recurrent caching \citep{zhang2025long,bulatov2023scaling}. For long output generation, methods like Suri \citep{pham2024surimulticonstraintinstructionfollowing} have explored multi-constraint instruction following, while LongWriter \citep{bai2024longwriter} introduced AgentWrite to enable ultra-long outputs by decomposing tasks into subtasks. Additionally, the Self-Lengthen framework \citep{quan2024language} iteratively expands initial outputs, training models to generate longer responses without requiring auxiliary data. These innovations enable LLMs to handle both long inputs and generate extended outputs, with parallel efforts focused on improving inference efficiency \citep{dumitru2025copyspec}.

\noindent \textbf{Long Generation Benchmarks}
Long generation benchmarks typically rely on similarity-based metrics like $\alpha$-nDCG and SelfBLEU, or LLM-as-a-Judge approaches~\cite{que2024hellobench, bai2024longwriter}, which struggle with longer texts due to their complexity. An alternative is to decompose evaluation into atomic statements, either extracted automatically using search engines or fixed databases for factual accuracy~\cite{song-etal-2024-veriscore, wei2024longform, samarinas2025beyond}, or manually designed through expert discussions~\cite{tan-etal-2024-proxyqa} or checklists~\cite{que2024hellobench}. However, these methods face verification challenges due to broad or trivial claims from automated extraction and incompleteness from manual design. To address this, objective tasks, such as MMLU \cite{liu-etal-2024-longgenbench} and procedural verification, provide more controlled evaluations but often misalign with real-world scenarios. While they support up to 4k tokens, they remain limited for longer texts, highlighting the need for more fine-grained and specialized benchmarks across different domains \citep{wang2025toward, yang2025llm}.

\subsection{Answer Length}
\label{sec:answer_length}

Our analysis of generated output lengths, detailed in Table~\ref{tab:avg_answer_length}, reveals discrepancies between instructed and actual token counts across all evaluated models. Key findings are as follows:
First, most models exhibit poor adherence to explicit length constraints, with the deviation increasing for longer targets (4k and 8k tokens). Second, distinct patterns emerge based on model type. Reasoning-oriented models (\modelname{DeepSeek-R1}) and certain proprietary models (\modelname{GPT-4o}) consistently produce outputs substantially shorter than requested. In contrast, the \modelname{Qwen3} series demonstrates more effective length control than its predecessor, \modelname{Qwen2.5}.
Crucially, we find no direct correlation between output length and overall task performance. Verbosity does not equate to higher quality, as many models that generate longer text achieve lower scores on our benchmark's core metrics.

\begin{table}[htbp]
\centering
\caption{Average output length for each model across four target length settings, highlighting models' adherence to length constraints.}
\label{tab:avg_answer_length}
\small
\resizebox{\linewidth}{!}{
\begin{tabular}{@{}lrrrr@{}}
\toprule
Model Name & 1k & 2k & 4k & 8k \\
\midrule
DeepSeek-R1-Distill-Qwen-32B & 874.2 & 1271.8 & 1404.6 & 1585.3 \\
DeepSeek-R1-Distill-Qwen-7B & 1430.5 & 1731.6 & 1931.0 & 1809.3 \\
LongWriter-glm4-9B & 1791.0 & 2156.1 & 2895.6 & 3277.4 \\
Meta-Llama-3-70B Instruct & 662.1 & 1504.6 & 2180.7 & 2613.9 \\
Meta-Llama-3-8B Instruct & 1155.8 & 1592.0 & 2119.5 & 2520.0 \\
Phi-4-mini-instruct & 2088.7 & 2594.3 & 2649.3 & 2606.7 \\
Qwen2.5-14B-Instruct & 1360.8 & 2188.8 & 2744.8 & 2987.2 \\
Qwen2.5-14B-Instruct (1M) & 913.2 & 1477.3 & 1678.4 & 2092.7 \\
Qwen2.5-3B Instruct & 1130.8 & 1558.7 & 1938.7 & 2178.6 \\
Qwen2.5-72B Instruct & 1242.9 & 1831.2 & 2341.0 & 2515.5 \\
Qwen2.5-7B Instruct & 1069.8 & 1669.7 & 2193.4 & 2351.4 \\
Qwen2.5-7B Instruct (1M) & 933.1 & 1370.8 & 1666.8 & 2087.3 \\
DeepSeek-R1 & 559.3 & 746.5 & 1078.0 & 1558.4 \\
SeepSeek-V3 & 674.7 & 1120.3 & 1405.2 & 1696.3 \\
Gemini-2.0-flash & 700.2 & 1139.4 & 1748.9 & 2385.3 \\
GPT-4o-2024-08-06 & 442.8 & 541.3 & 636.4 & 739.4 \\
GPT-4o-2024-11-20 & 649.5 & 802.6 & 917.0 & 947.9 \\
Llama-4-maverick-17b-128e-instruct & 779.1 & 1160.0 & 1331.5 & 1780.2 \\
Llama-4-scout-17b-16e-instruct & 953.6 & 1484.9 & 1888.5 & 2318.0 \\
o3-mini-2025-01-31 & 1057.0 & 2136.4 & 2958.8 & 2948.1 \\
Qwen-long & 647.4 & 952.8 & 1364.4 & 1733.8 \\
Qwen3-14b & 727.8 & 2208.0 & 2836.7 & 3750.7 \\
Qwen3-32b & 655.3 & 1354.9 & 2411.2 & 3278.3 \\
Qwen3-4b & 658.5 & 1436.0 & 2455.8 & 3227.5 \\
Qwen3-8b & 670.2 & 1552.8 & 2666.7 & 3635.0 \\
QwQ-plus-2025-03-05 & 606.0 & 857.4 & 1263.6 & 1654.2 \\
\bottomrule

\end{tabular}}
\end{table}

\begin{table*}[htbp]
  \centering
  \caption{Configuration of large language models, their backends and decoding parameters provided by different suppliers}
  \footnotesize
  \label{tab:model_parameters}
  \begin{tabularx}{\textwidth}{ll >{\RaggedRight}X >{\RaggedRight\ttfamily}X}
    \toprule
    \textbf{Provider} & \textbf{Backend} & \textbf{Model} & \textbf{Decoding Parameters} \\
    \midrule

    \multirow{3}{*}{Meta} 
    & vLLM & Meta-Llama-3.1-8B-Instruct / Meta-Llama-3.1-70B-Instruct & temperature: 0.7, top\_p: 0.8, max\_tokens: 8192, stream: False \\
    \cmidrule(l){2-4}
    & \multirow{2}{*}{Aliyun Dashscope} & llama-4-scout-17b-16e-instruct & temperature: 0.7, max\_tokens: 8192, stream: True \\
    & & llama-4-maverick-17b-128e-instruct & temperature: 0.7, max\_tokens: 8192, stream: True \\
    \midrule

    \multirow{3}{*}{OpenAI}
    & \multirow{3}{*}{OpenAI API} & gpt-4o-2024-11-20 & temperature: 0.7, max\_tokens: 8192, stream: True \\
    & & o3-mini-2025-01-31 & temperature: 0.7, max\_tokens: 8192, stream: True \\
    & & gpt-4o-mini-2024-07-18 & temperature: 0.7, max\_tokens: 8192, stream: True \\
    \midrule

    \multirow{2}{*}{Google}
    & OpenAI API & gemini-2.0-flash & temperature: 0.7, max\_tokens: 8192, stream: True \\
    \cmidrule(l){2-4}
    & vLLM & gemma-3-12b-it / gemma-3-27b-it & temperature: 0.7, max\_tokens: 8192, stream: False \\
    \midrule

    Zhipu AI & vLLM & LongWriter-glm4-9b & max\_tokens: 8192, stream: False \\
    \midrule

    Microsoft & vLLM & Phi-4-mini-instruct & max\_tokens: 8192, stream: False \\
    \midrule

    \multirow{2}{*}{DeepSeek}
    & Aliyun Dashscope & deepseek-v3 / deepseek-r1 & temperature: 0.7, max\_tokens: 8192, stream: True \\
    \cmidrule(l){2-4}
    & vLLM & DeepSeek-R1-Distill-Qwen-32B & max\_tokens: 8192, stream: False \\
    \midrule

    \multirow{8}{*}{Alibaba}
    & vLLM & Qwen2.5-14B-Instruct-1M & max\_tokens: 8192, stream: False \\
    \cmidrule(l){2-4}
    & Aliyun Dashscope & qwen-long & temperature: 0.7, max\_tokens: 8192, stream: True \\
    \cmidrule(l){2-4}
    & vLLM & Qwen2.5-3B/7B/14B/72B-Instruct & temperature: 0.7, top\_p: 0.8, max\_tokens: 8192, stream: False \\
    \cmidrule(l){2-4}
    & vLLM & qwen3-14b-r & temperature: 0.7, top\_p: 0.8, max\_tokens: 32768, stream: False \\
    \cmidrule(l){2-4}
    & Aliyun Dashscope & qwen3-14b & temperature: 0.7, top\_p: 0.8, max\_tokens: 8192 \\
    \cmidrule(l){2-4}
    & vLLM & qwen3-32b & temperature: 0.7, top\_p: 0.8, max\_tokens: 8192, stream: False \\
    \cmidrule(l){2-4}
    & Aliyun Dashscope & qwen3-8b / qwen3-14b / qwen3-32b & temperature: 0.7, max\_tokens: 8192, stream: True \\
    \cmidrule(l){2-4}
    & Aliyun Dashscope & QwQ-plus / qwen-max series & temperature: 0.7, top\_p: 0.8, presence\_penalty: 1.5, max\_tokens: 8192, stream: True \\

    \bottomrule
  \end{tabularx}
\end{table*}

\subsection{AP Style Criteria}
\begin{figure*}[htbp]
\centering
\includegraphics[width=0.9\linewidth, keepaspectratio]{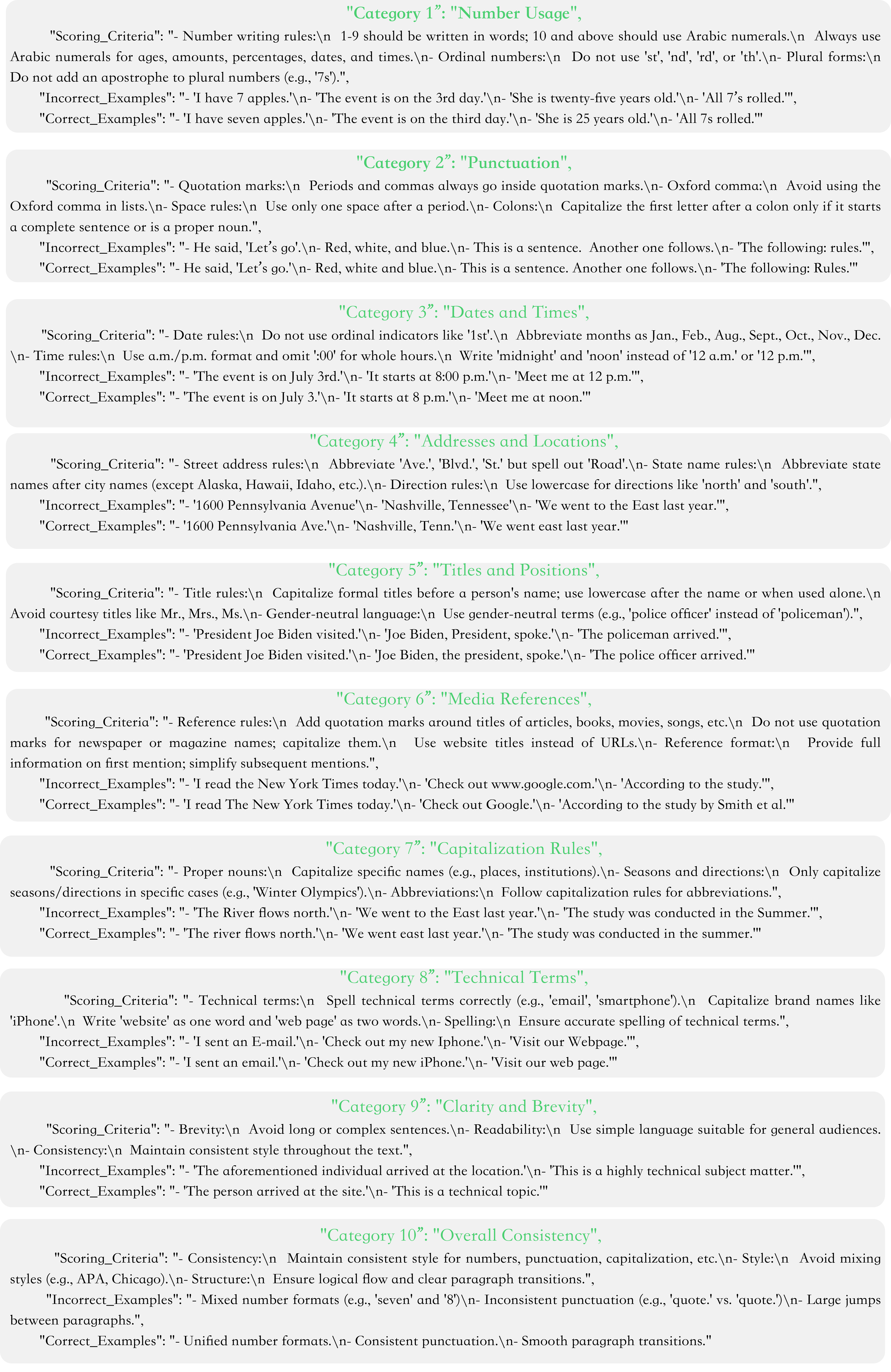}
\caption{Generation and Evaluation Rules for Constraints in the News Writing Task. Ten dimensions, each containing rules, positive examples, and error cases that guided the creation of verifiable test instances.}
\label{fig:ap_criteria}
\end{figure*}

In our AP Style News Writing task, we assess a model's proficiency in adhering to complex, real-world stylistic guidelines from the Associated Press (AP) Stylebook. To ensure an objective and verifiable evaluation, we moved beyond holistic review and focused on verifiable rules.
The creation of our test cases was guided by ten distinct categories of AP Style rules, as detailed in Figure~\ref{fig:ap_criteria}. For each category, we generated Constraint-Verifier pairs where the Constraint is a factual statement deliberately crafted to violate a specific rule (e.g., writing "7 apples" instead of "seven apples"). The corresponding Verifier is the same statement, corrected to be fully compliant with AP style. The model is then tasked with incorporating the factual information from the incorrect Constraint into its generated article, but in the stylistically correct form.

\renewcommand{\arraystretch}{1.2}
\newcolumntype{Y}{>{\raggedright\arraybackslash}X}
\definecolor{headerbg}{RGB}{234,242,248}
\definecolor{rulecolor}{RGB}{217,217,217}

\subsection{Details of sample construction}
\noindent \textbf{CSV Sales Report Analysis (SR):} The methodology facilitates the generation of synthetic transactional sales data intrinsically correlated with corresponding analytical conclusions. The process commences with the definition of foundational sales scenario parameters, including sales region, target fiscal period, currency, overarching sales targets, and antecedent period sales figures, alongside a predefined corpus of sales representatives, product lines, and operational cities. A pivotal aspect involves the stochastic injection of predefined systemic biases during each operational instance; these biases may pertain to overall target achievement (e.g., exceeding, meeting, or missing targets), growth trajectory (positive, neutral, or negative), the anomalous performance of specific sales representatives or products, and variations in new customer acquisition rates.
These stochastically determined biases subsequently modulate the synthesis of individual transactional records. Attributes of each transaction, such as sales representative assignment, product selection, customer provenance (new versus existing), and critically, the final transaction value, are probabilistically influenced by the afore-mentioned biases. This ensures that the generated dataset not only achieves a specified volume but also exhibits inherent, bias-driven characteristics across multiple dimensions, thereby providing a feature-rich foundation for subsequent analytical procedures.
Upon completion of data synthesis, the resultant structured dataset is subjected to a multi-dimensional analytical engine. This engine emulates real-world business intelligence reporting by performing comprehensive quantitative aggregations and inferential processing across diverse facets, including overall performance metrics (e.g., total sales versus target, period-over-period growth, average transaction value), sales representative efficacy (e.g., top and bottom performers, target attainment distributions), product performance (e.g., leading revenue generators, category contributions), geographical sales distribution, and customer segment analysis (e.g., new versus existing customer value, key account contributions).
Key metrics and identified trends derived from this analysis are then articulated as concise, natural language analytical conclusions. To enhance utility and stimulate further inquiry, each conclusion is systematically paired with a relevant analytical query, designed to prompt deeper investigation into the causal factors underpinning the observed phenomena. The system culminates in the delivery of two principal outputs: the raw, granular transactional dataset (typically in CSV format), which serves as the evidentiary basis for analysis, and a structured compendium (typically in JSON format) containing metadata, key performance indicators, and a curated, prioritized set of "conclusion-query" pairings, offering directly consumable insights for simulated business reporting. This integrated pipeline underscores a design philosophy centered on the coherent synthesis of data with its analytical interpretation.

\noindent \textbf{Code Fixing with Flake8 Compliance (CF):} The system employs a generative methodology to synthesize Python source code exhibiting a high density of nuanced linting violations, intended to serve as challenging test instances for static analysis tools and code quality assessment. The generation process is initiated by establishing global configuration parameters, including stylistic targets (e.g., line length) and complexity constraints (e.g., maximum nesting depth, function length), alongside lexical resources such as curated lists of nouns, verbs, and adjectives for constructing semantically plausible, albeit potentially misleading, identifiers.
A core component is a dynamic scope management system, which tracks variable definitions and usage across nested lexical contexts. This enables the generation of syntactically valid code where identifier-related violations, such as improper naming conventions (e.g., N-series violations from flake8-naming) or unused variables (F841), are contextually embedded. Identifier generation itself is a probabilistic process, designed to stochastically introduce deviations from Python Enhancement Proposal 8 (PEP 8) style guidelines, while also attempting to create names that might subtly obscure their true purpose or shadow existing identifiers in parent scopes.
The synthesis of executable code blocks and function bodies is orchestrated through a weighted, probabilistic selection of diverse code constructs. These constructs range from simple assignments and print statements to complex control flow structures like conditional statements and loops. Each construct generator is imbued with the capability to introduce specific categories of violations. For instance, conditional statement generators might create explicit boolean comparisons (SIM21x) or if-else patterns amenable to ternary expressions (SIM108). Loop generators may produce unconventional iterator variable names or inefficient comprehensions (C4xx series from flake8-comprehensions). Furthermore, generators for function definitions are specifically designed to introduce more complex issues, such as mutable default arguments (B006 from flake8-bugbear) or function calls within default argument expressions (B008), often obfuscated by the presence of other parameters and non-trivial function bodies.
Whitespace and formatting violations (E-series and W-series) are pervasively introduced at various granularities, from inconsistent spacing around operators and after commas to improper blank line usage and trailing whitespace. The system also synthesizes a sequence of inter-dependent functions, simulating a rudimentary program flow (e.g., data loading, validation, analysis, reporting), which are ultimately orchestrated within a main execution block. This structural coherence provides a more realistic backdrop for the embedded violations, moving beyond isolated infractions to scenarios requiring more holistic refactoring. The overall probability of introducing a violation is a configurable parameter, allowing for control over the density of infractions, with the system actively aiming to make these violations less trivial to automatically or manually remediate by intertwining them with functional, albeit flawed, program logic. The final output is a runnable Python script, replete with these intentionally challenging, multi-category linting issues.

\noindent \textbf{KG to Text Biography Generation (BioG):} system synthesizes rich, protagonist-centric knowledge graphs (KGs) and subsequently translates salient subgraphs into natural language narratives. The generative process for each KG commences with the instantiation of a unique protagonist, whose attributes, including socio-economic background and a randomly assigned character archetype (e.g., Scientist, Artist, Entrepreneur), are stochastically determined. These initial conditions significantly influence the subsequent probabilistic expansion of the KG. The protagonist's lifespan and historical era are also established to ensure temporal coherence for related entities and events.
The KG is then incrementally constructed through an iterative expansion process originating from the protagonist. At each step, existing nodes are selected for expansion based on their proximity to the protagonist and predefined archetypal relationship propensities. New nodes, representing persons, organizations, places, creative works, or events, are generated with contextually relevant attributes, or existing nodes are connected, adhering to a set of permissible relationship types defined within a structured map. This map also dictates the likelihood of specific relationships based on the source node's type and, for persons, their current life phase (e.g., Childhood, Education, MidCareer). Attribute generation for new entities, such as names, job titles, or event descriptions, leverages procedural generation techniques and controlled randomness, often influenced by the protagonist's established background and archetype to foster narrative consistency. Temporal plausibility is rigorously maintained by ensuring that dates associated with relationships and events align with the lifespans of involved entities.
Once a KG reaches a target size or expansion limits, a focused subgraph is extracted. This subgraph typically comprises nodes within a specified graph distance from the protagonist, representing the most narratively relevant portion of the larger KG. This subgraph then serves as the direct input for the text generation phase. Each node attribute (excluding the primary name) and each relationship within this subgraph, along with significant attributes of these relationships (e.g., roles, dates, specific details like degree or investment amount), are systematically converted into individual descriptive sentences using predefined, templated linguistic patterns. These patterns map structural KG elements (subject-predicate-object triples, or subject-attribute-value) to natural language constructs.
The system's output for each generated KG is multi-faceted, including the full KG data, the extracted subgraph data, and the derived natural language sentences, typically stored in structured JSON files. Optionally, visualizations of both the full KG and the subgraph can be produced using graph layout algorithms. Finally, as an aggregative step, the natural language sentences generated from all individual KGs within a single execution run are compiled into a consolidated dataset, facilitating larger-scale analysis or downstream natural language processing tasks. This methodology emphasizes the creation of datasets where structured knowledge and its textual manifestation are coherently and traceably linked, grounded in simulated sociological and temporal contexts.

\noindent \textbf{AP Style News Writing (NW):} The system under discussion is designed to rigorously evaluate a large language model's (LLM) proficiency in generating news reports that conform to the Associated Press (AP) style guidelines. This evaluation is predicated on the model's ability to synthesize a coherent narrative based on a given news topic query, integrate a series of predefined factual statements, and adhere to a specified target word count, all while meticulously applying AP style conventions.
The process of generating verifiable test data, specifically the factual statements, is a critical precursor to the evaluation. These statements are meticulously crafted to serve as direct inputs that the LLM must incorporate into its generated news article. Crucially, each statement is designed to test a specific facet of the AP style guide; thus, many are intentionally formulated to violate these rules. For instance, a statement might employ incorrect number usage (e.g., writing out "eleven" instead of using the numeral "11"), misuse punctuation (e.g., including an Oxford comma), or improperly format dates, times, or titles. Accompanying each such potentially flawed statement in the test dataset is its corresponding correct AP style expression and a clear rationale explaining the nature of the original stylistic error. This structured approach ensures that each statement serves as a verifiable unit for assessing the LLM's capacity for rule-based stylistic correction.
The construction of the prompt provided to the generative LLM is a multi-component process. It begins with the query, which defines the overarching news topic, often suggesting a narrative structure or specific angles to be explored. To this, the complete AP style rubric—a comprehensive guide detailing rules across numerous categories with illustrative examples—is appended. A key element of the prompt is a curated list of the aforementioned factual statements. These statements, presented in their original, potentially non-compliant form, are explicitly designated as mandatory inclusions for the generated article. The prompt also specifies the target word count, imposing a length constraint on the LLM's output. This careful assembly of the prompt creates a challenging scenario where the LLM must not only generate fluent and relevant content based on the query but also actively engage with the AP style guide to identify and rectify the stylistic infelicities within the provided statements as they are woven into the narrative. The verifiability of the task lies in the direct comparison of the model's treatment of these embedded statements against their known correct AP style forms, all within the context of the broader news writing assignment.

\subsection{Evaluation Efficiency}
\label{sec:efficiency}

Given the scale of our evaluation (23 models across 5,600 samples), understanding these efficiency aspects is crucial.

For this analysis, we utilized two nodes, each equipped with 8 NVIDIA A100 GPUs (totaling 16). One node was dedicated to deploying the evaluator model, Qwen2.5 72B, while the other hosted the model under test, Llama3.1 8B. To establish a clear baseline, we measured performance in a single-threaded mode. We recorded the inference and evaluation times for a single sample across seven distinct tasks, varying the input context length at four levels: 1k, 2k, 4k, and 8k tokens.

The results of this single-threaded performance analysis are presented in Table~\ref{tab:efficiency_analysis}. As shown, the evaluation time can be a significant component of the total processing time, particularly for tasks like News Writing (NW), which involve complex rubric-based assessments. For tasks such as Knowledge Graph (KVG) and Sales Message (SMS), the evaluation is nearly instantaneous as it relies on simple keyword matching. The full names for the task abbreviations can be found in Table~2 of the main paper.

\begin{table*}[htbp]
\centering
\caption{Efficiency analysis of the evaluation pipeline. Each value represents the time in seconds to process a single sample in single-threaded mode. The analysis was conducted using Qwen2.5 72B as the evaluator and Llama3.1 8B as the model being tested.}
\label{tab:efficiency_analysis}
\begin{tabular}{lrrrrrrr}
\toprule
\textbf{Metric (Input Length)} & \textbf{CF} & \textbf{BioG} & \textbf{SR} & \textbf{NW} & \textbf{KVG} & \textbf{SMS} & \textbf{PR} \\
\midrule
Inference (1k) & 29.72 & 20.58 & 16.29 & 45.10 & 139.20 & 134.95 & 35.99 \\
Evaluation (1k) & 7.31 & 9.38 & 4.54 & 87.04 & 0.00 & 0.00 & 0.01 \\
\midrule
Inference (2k) & 28.86 & 16.49 & 24.63 & 22.44 & 129.35 & 133.46 & 23.75 \\
Evaluation (2k) & 5.34 & 2.42 & 8.89 & 120.05 & 0.00 & 0.00 & 0.00 \\
\midrule
Inference (4k) & 41.12 & 121.53 & 16.04 & 32.95 & 161.03 & 146.42 & 31.42 \\
Evaluation (4k) & 2.54 & 20.13 & 12.17 & 155.70 & 0.00 & 0.00 & 0.01 \\
\midrule
Inference (8k) & 106.56 & 33.62 & 36.04 & 32.91 & 193.31 & 149.92 & 90.09 \\
Evaluation (8k) & 5.49 & 30.32 & 12.96 & 1439.32 & 0.00 & 0.00 & 0.01 \\
\bottomrule
\end{tabular}
\end{table*}

To optimize efficiency in our main experiments, we employed parallel processing. For inference, we used 16 parallel threads. For the evaluation stage, we utilized 5 parallel threads with appropriate batch sizes tailored to the task (e.g., a batch size of 5 for Sales Report and Knowledge Graph tasks, and 10 for AP Style News). The evaluation pipeline ran on eight A100 GPUs, while inference was performed either on a separate set of eight A100 GPUs for local models or via API calls for proprietary models. This parallelized setup significantly reduced the overall wall-clock time required for our large-scale benchmark.

\subsection{Additional Data Examples}
\begin{figure*}[htbp]
\centering
\includegraphics[width=\linewidth, keepaspectratio]{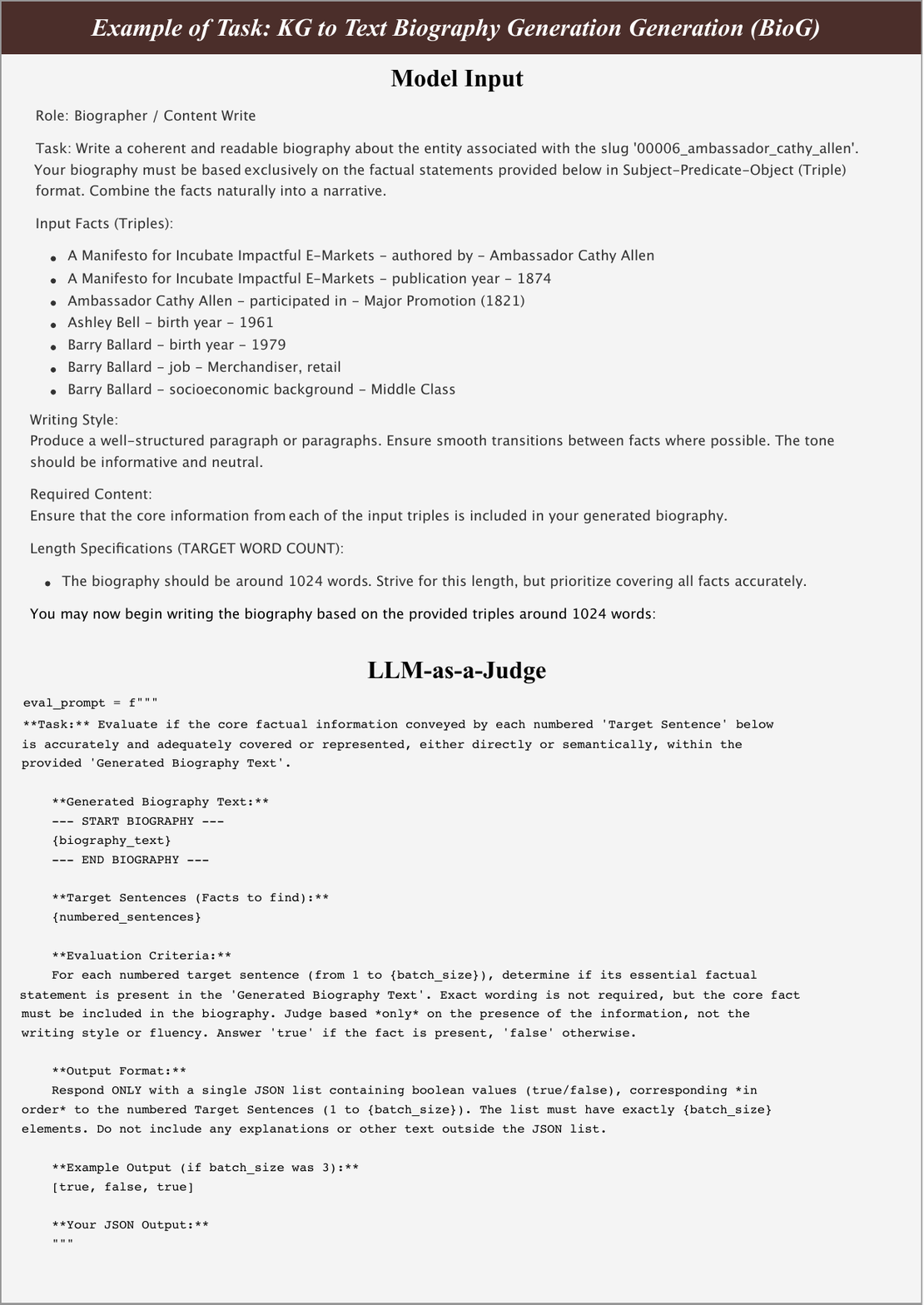}
\caption{An illustrative example for the KG to Text Biography Generation (BioG) task.}
\label{fig:example_biog}
\end{figure*}

\begin{figure*}[htbp]
\centering
\includegraphics[width=\linewidth, keepaspectratio]{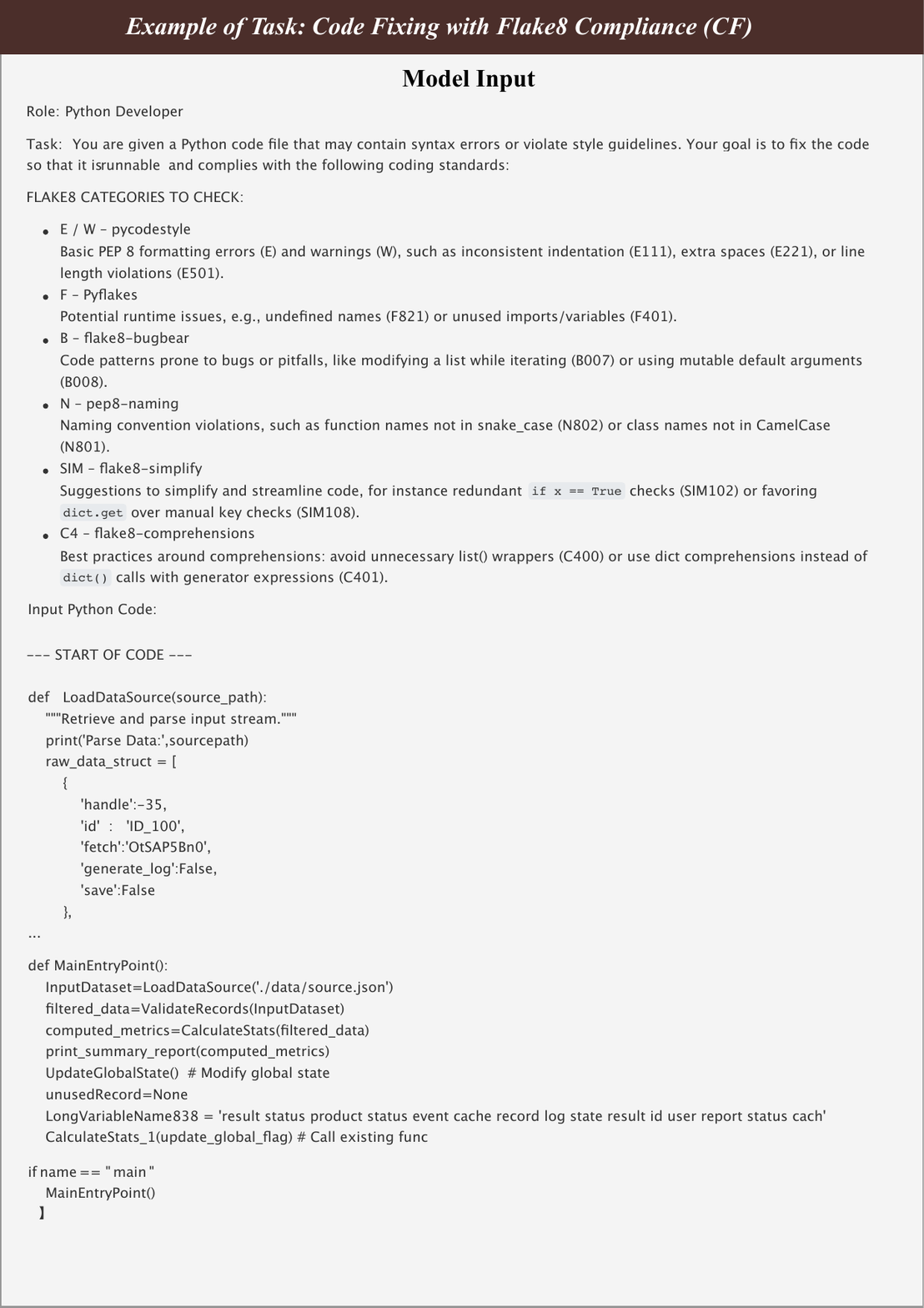}
\caption{An illustrative example for the Code Fixing (CF) task (Part 1/2).}
\label{fig:example_cf_part1}
\end{figure*}

\begin{figure*}[htbp]
\centering
\includegraphics[width=\linewidth, keepaspectratio]{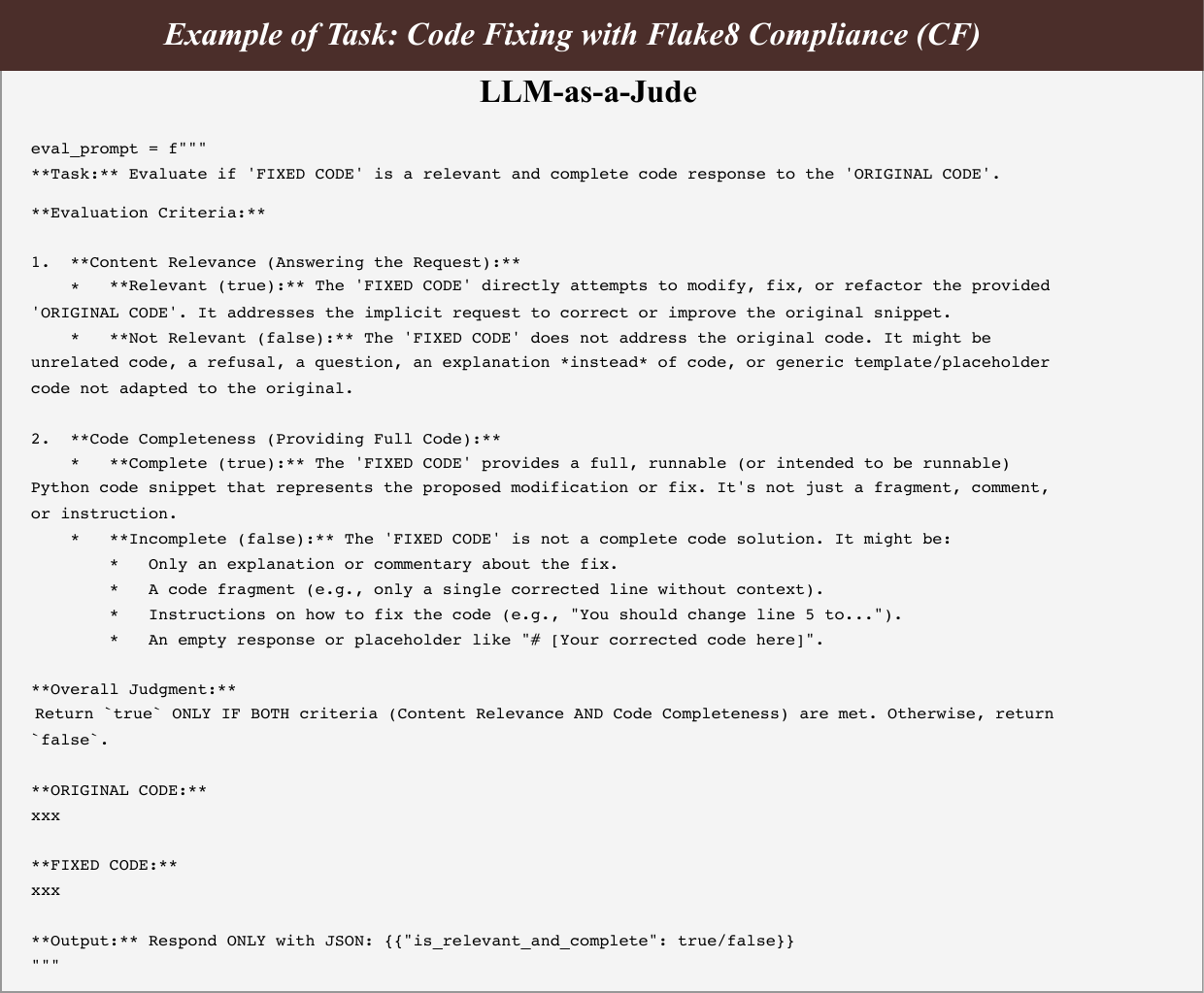}
\caption{An illustrative example for the Code Fixing (CF) task (Part 2/2).}
\label{fig:example_cf_part2}
\end{figure*}

\begin{figure*}[htbp]
\centering
\includegraphics[width=\linewidth, keepaspectratio]{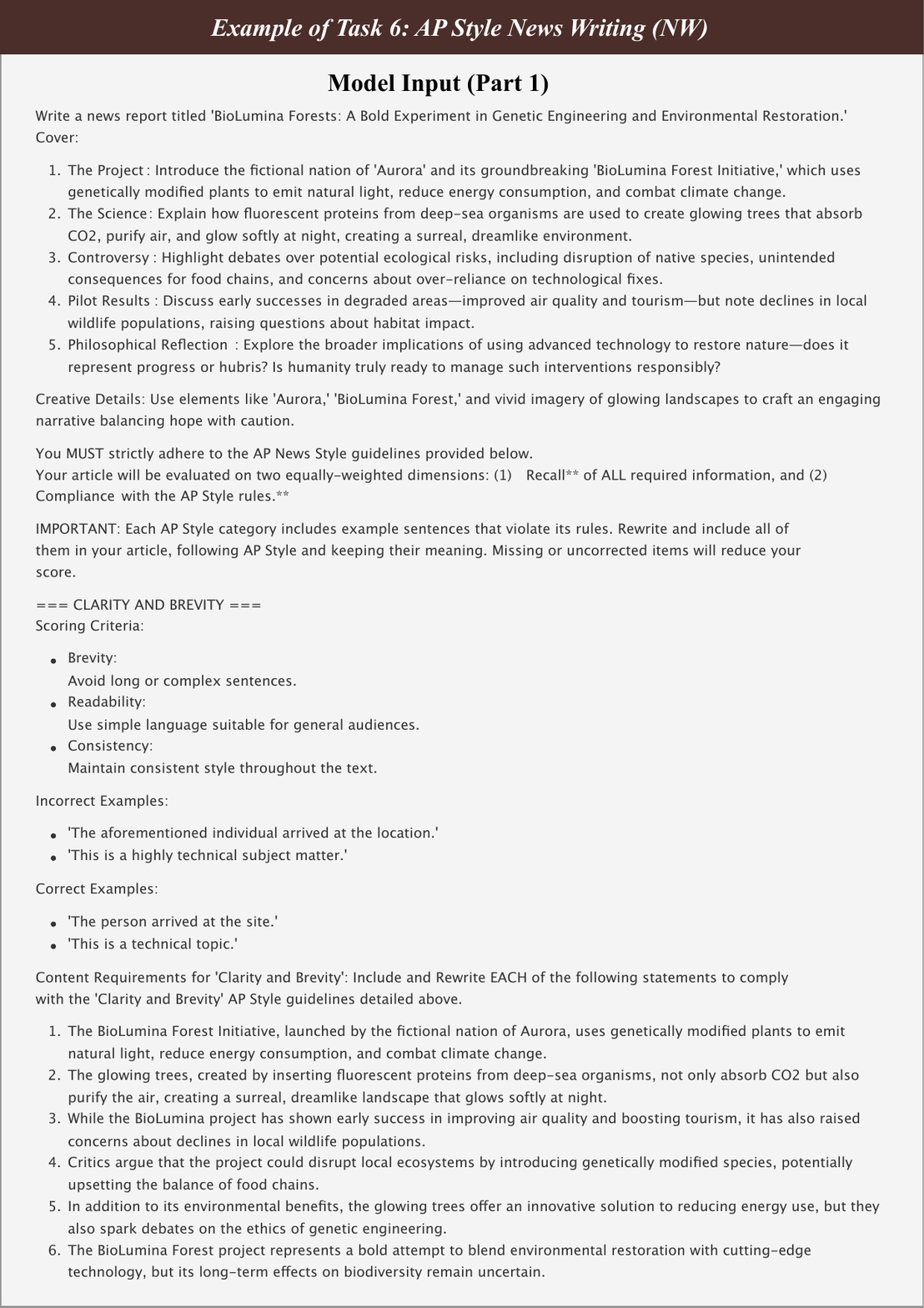}
\caption{An illustrative example for the AP Style News Writing (NW) task (Part 1/3).}
\label{fig:example_nw_part1}
\end{figure*}

\begin{figure*}[htbp]
\centering
\includegraphics[width=\linewidth, keepaspectratio]{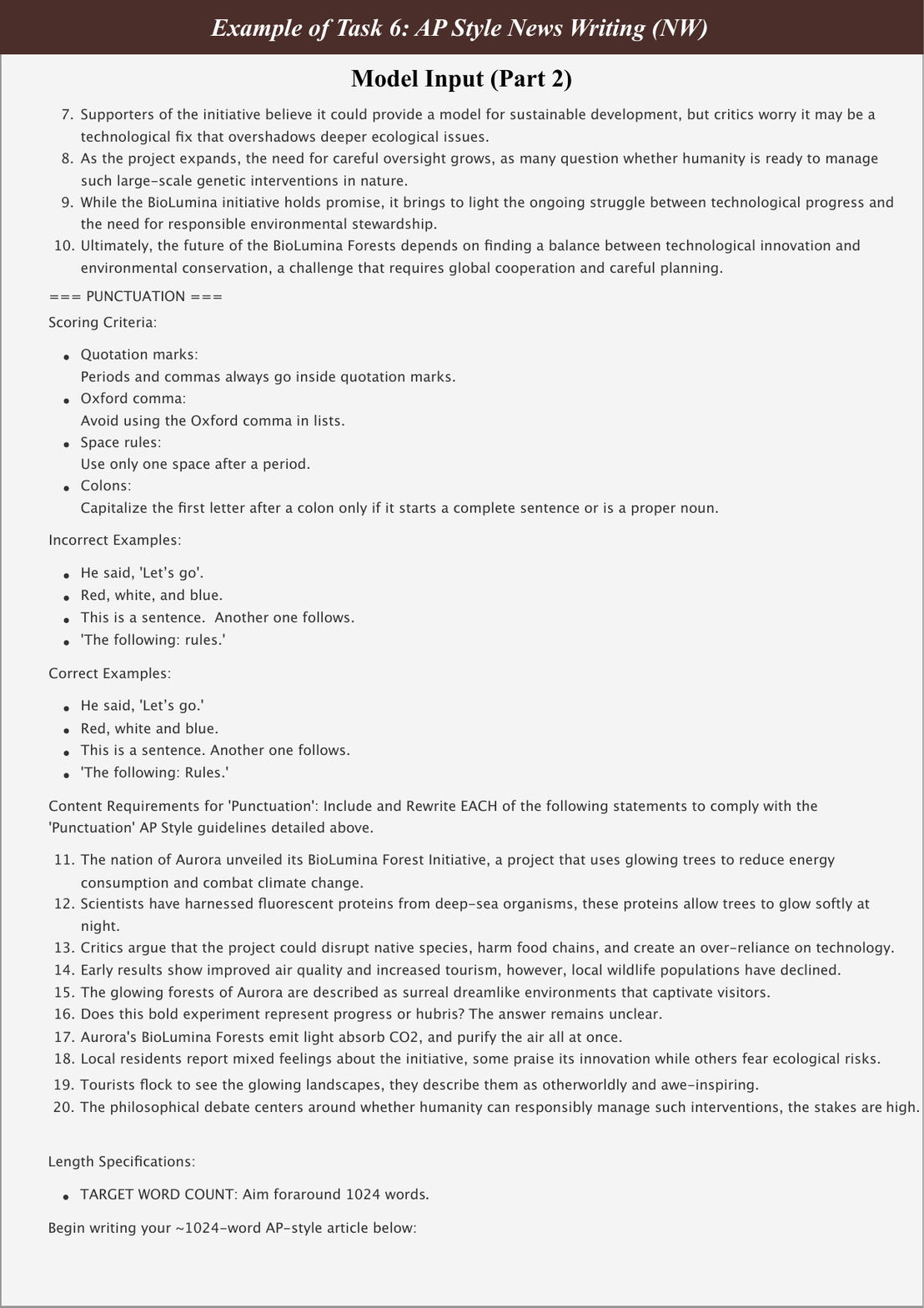}
\caption{An illustrative example for the AP Style News Writing (NW) task (Part 2/3).}
\label{fig:example_nw_part2}
\end{figure*}

\begin{figure*}[htbp]
\centering
\includegraphics[width=\linewidth, keepaspectratio]{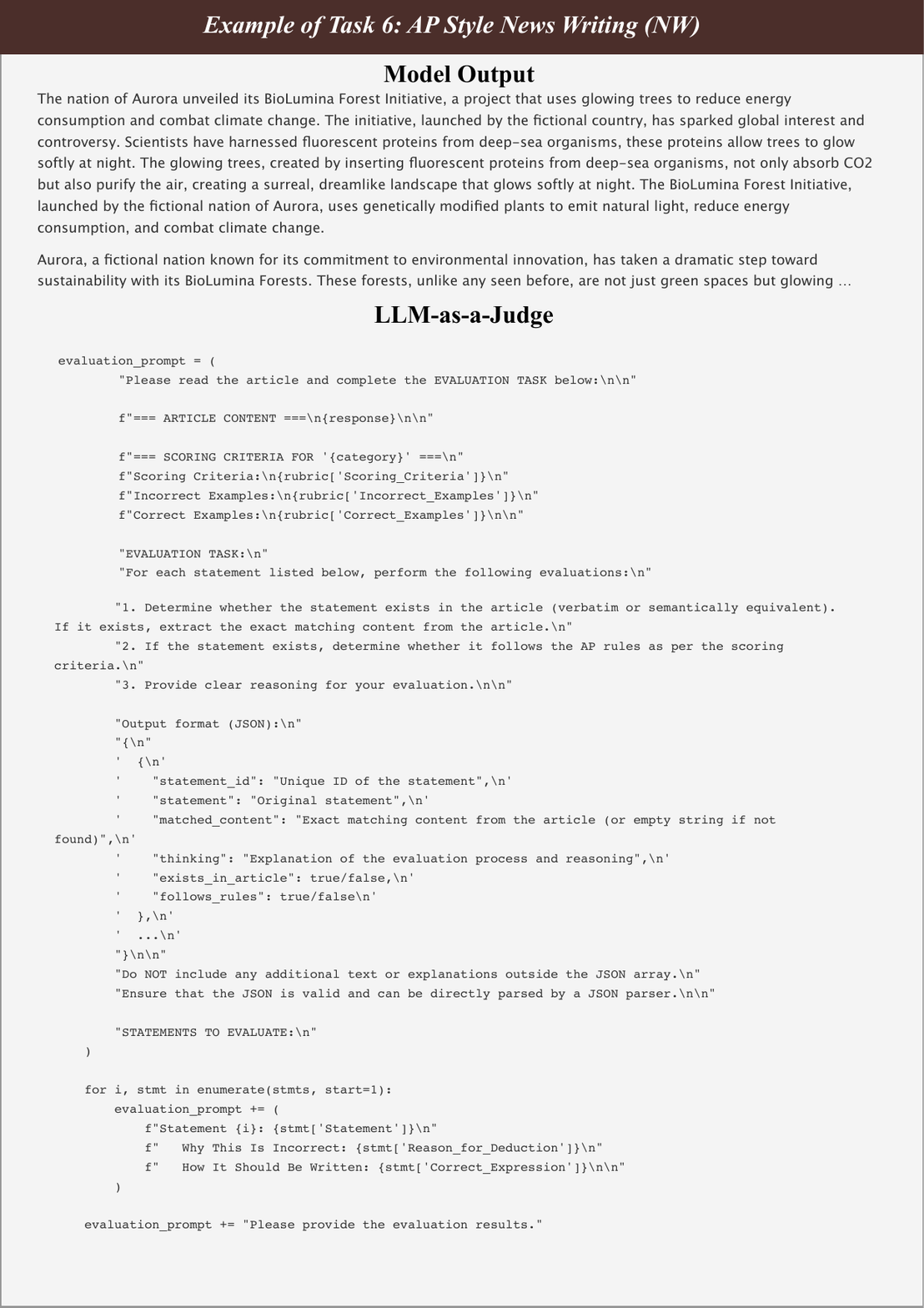}
\caption{An illustrative example for the AP Style News Writing (NW) task (Part 3/3).}
\label{fig:example_nw_part3}
\end{figure*}

\begin{figure*}[htbp]
\centering
\includegraphics[width=\linewidth, keepaspectratio]{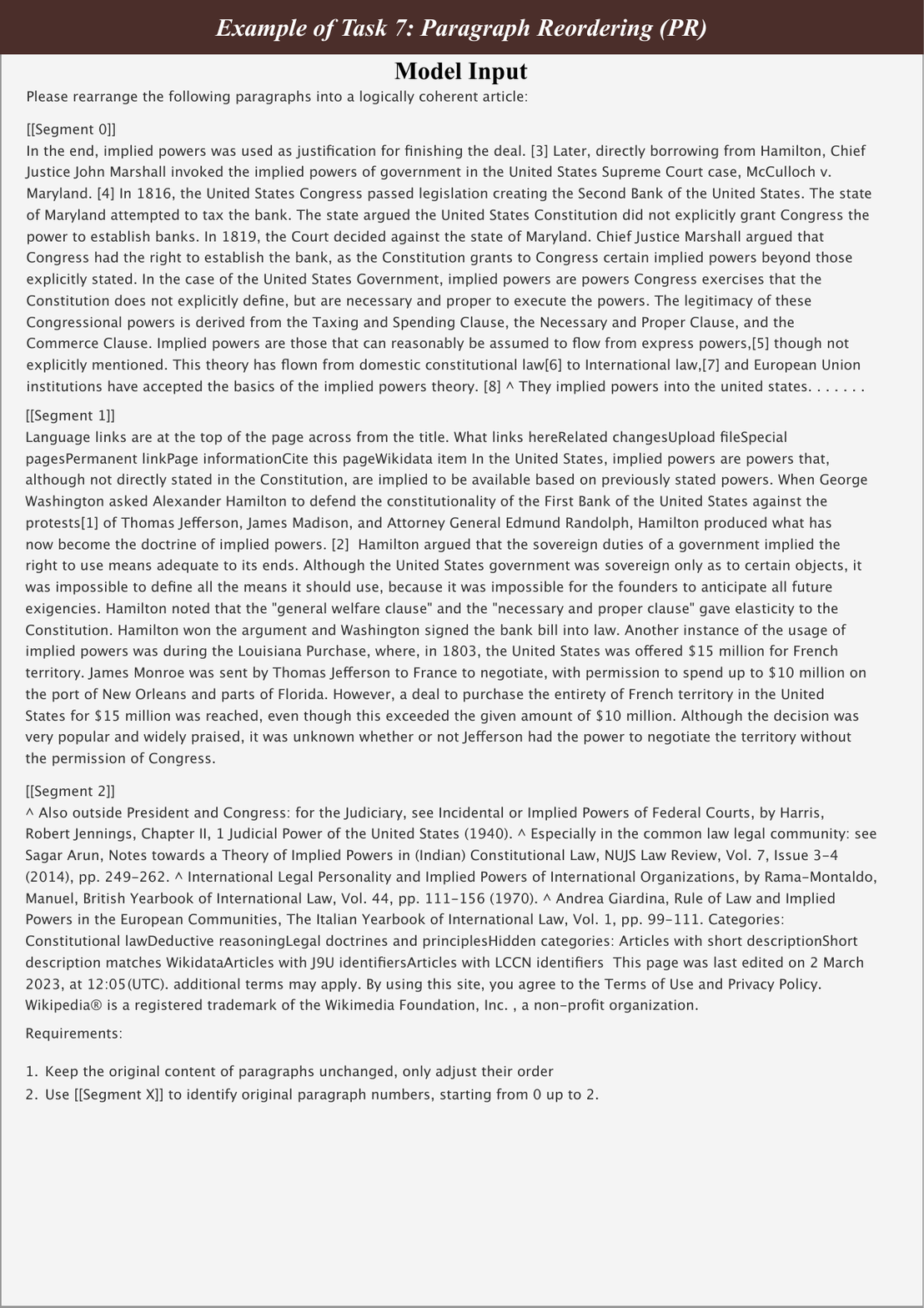}
\caption{An illustrative example for the Paragraph Reordering (PR) task.}
\label{fig:example_pr}
\end{figure*}

\begin{figure*}[htbp]
\centering
\includegraphics[width=\linewidth, keepaspectratio]{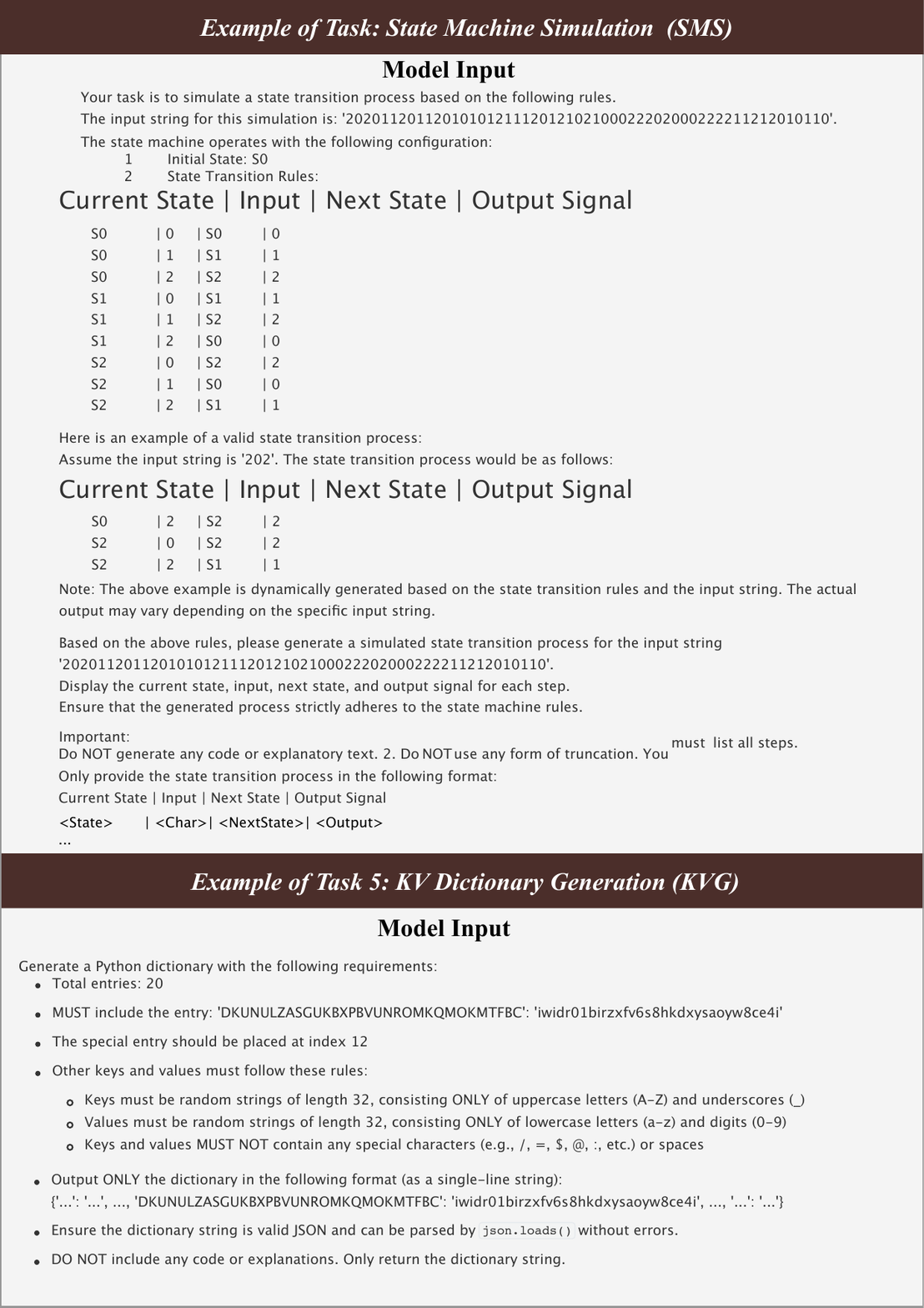}
\caption{Illustrative examples for the State Machine Simulation (SMS) and KV Dictionary Generation (KVG).}
\label{fig:example_sms_kvg}
\end{figure*}

\begin{figure*}[htbp]
\centering
\includegraphics[width=\linewidth, keepaspectratio]{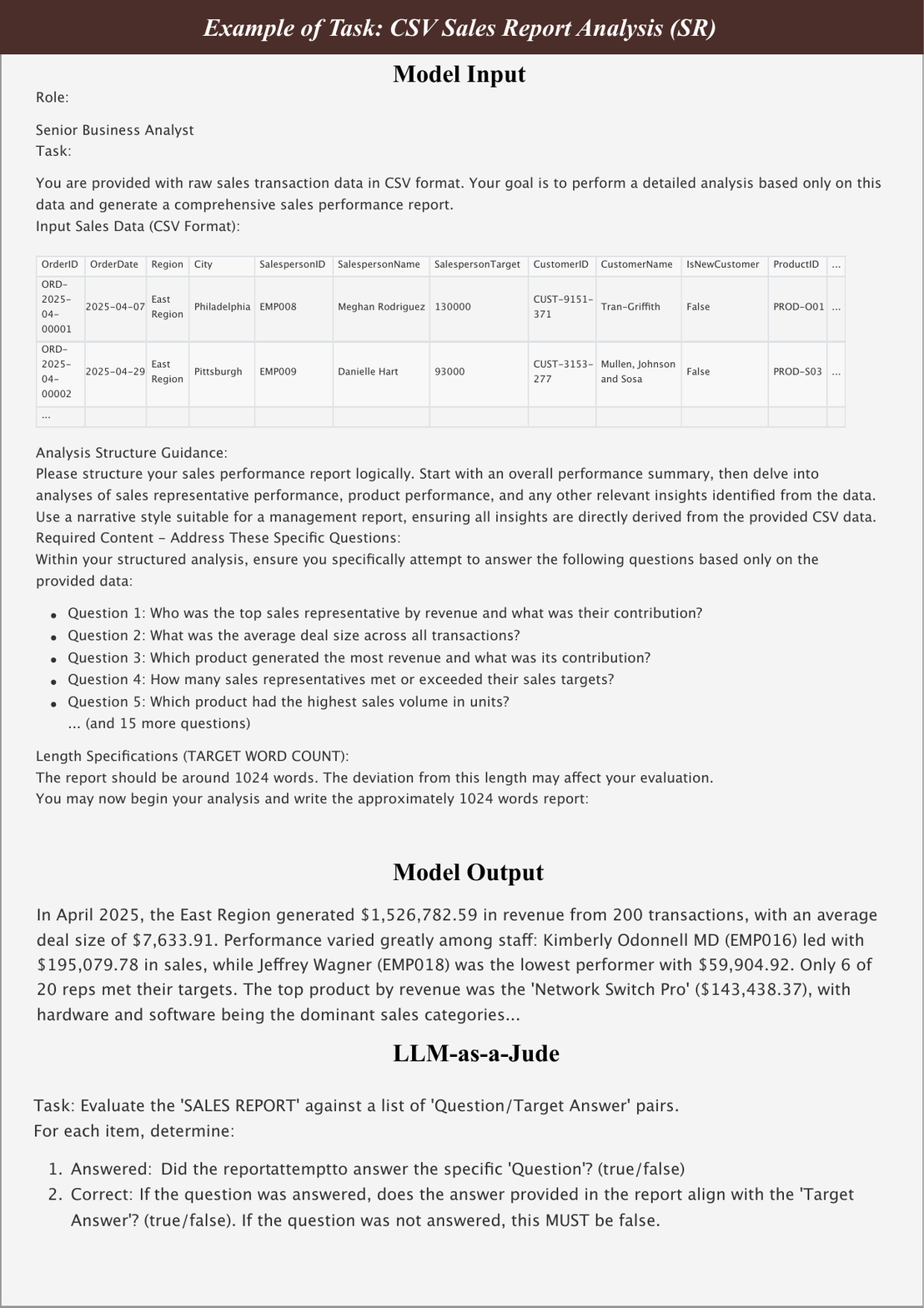}
\caption{An illustrative example for the CSV Sales Report Analysis (SR) task (Part 1/2).}
\label{fig:example_sr_part1}
\end{figure*}

\begin{figure*}[htbp]
\centering
\includegraphics[width=\linewidth, keepaspectratio]{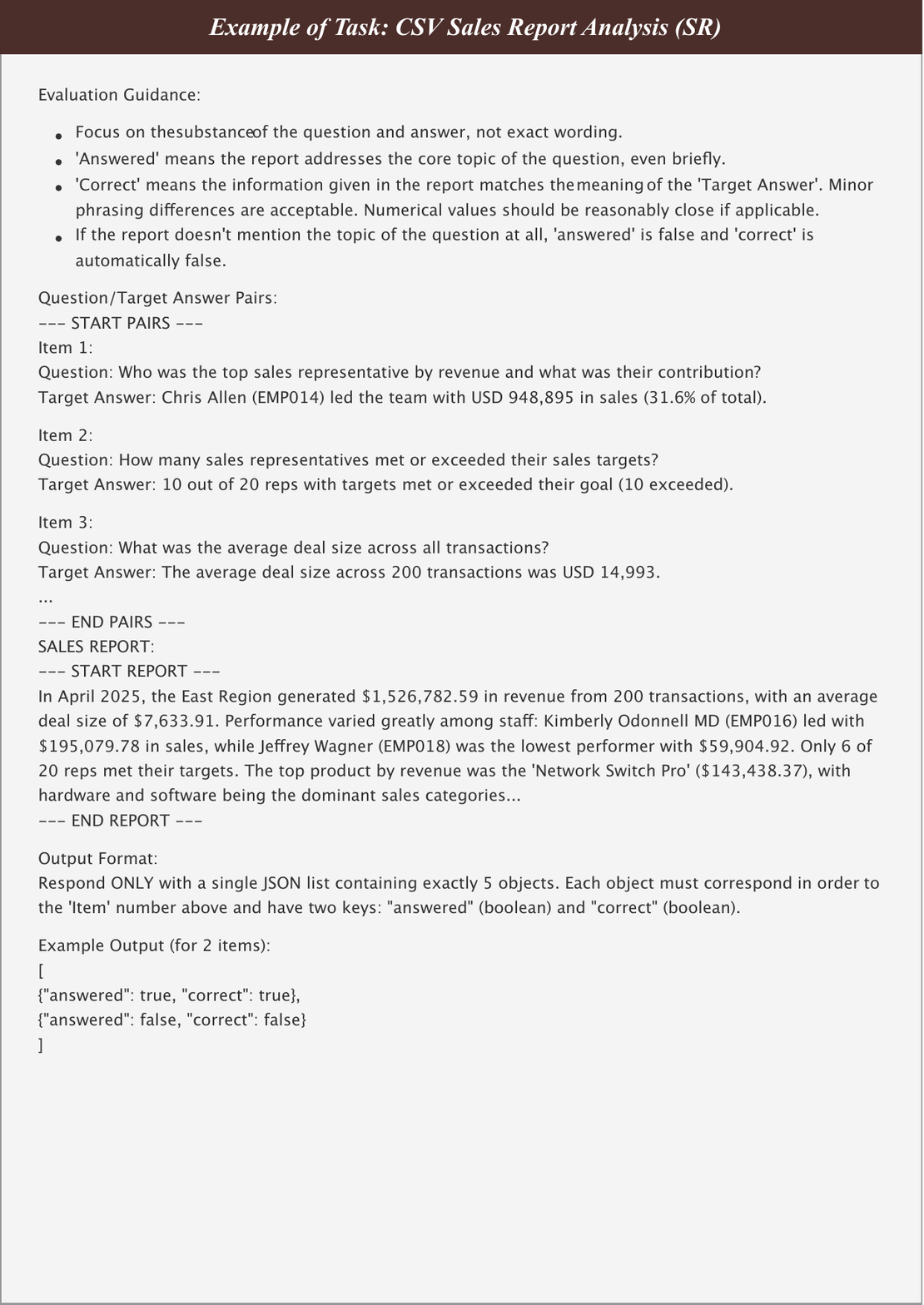}
\caption{An illustrative example for the CSV Sales Report Analysis (SR) task (Part 2/2).}
\label{fig:example_sr_part2}
\end{figure*}

To provide a concrete understanding of the tasks within the LongWeave benchmark, this section presents illustrative examples of the input prompts used during evaluation. These examples showcase the structure of the input materials, the detailed instructions, and the specific constraints that models must follow. The figures below cover all seven tasks: KG to Text Biography Generation (Figure~\ref{fig:example_biog}); Code Fixing (Figure~\ref{fig:example_cf_part1} and~\ref{fig:example_cf_part2}); AP Style News Writing, which includes the topic, factual statements, and style rubric (Figure~\ref{fig:example_nw_part1},\ref{fig:example_nw_part2}, and\ref{fig:example_nw_part3}); Paragraph Reordering (Figure~\ref{fig:example_pr}); State Machine Simulation and KV Dictionary Generation (Figure~\ref{fig:example_sms_kvg}); and CSV Sales Report Analysis, which details the data and questions (Figure~\ref{fig:example_sr_part1} and~\ref{fig:example_sr_part2}). Collectively, these examples demonstrate the diversity of challenges posed by LongWeave and provide insight into the practical implementation of CoV-Eval.

\subsection{Details of Task Generator}

\begin{figure*}[htbp]
    \centering
    \includegraphics[width=1\textwidth]{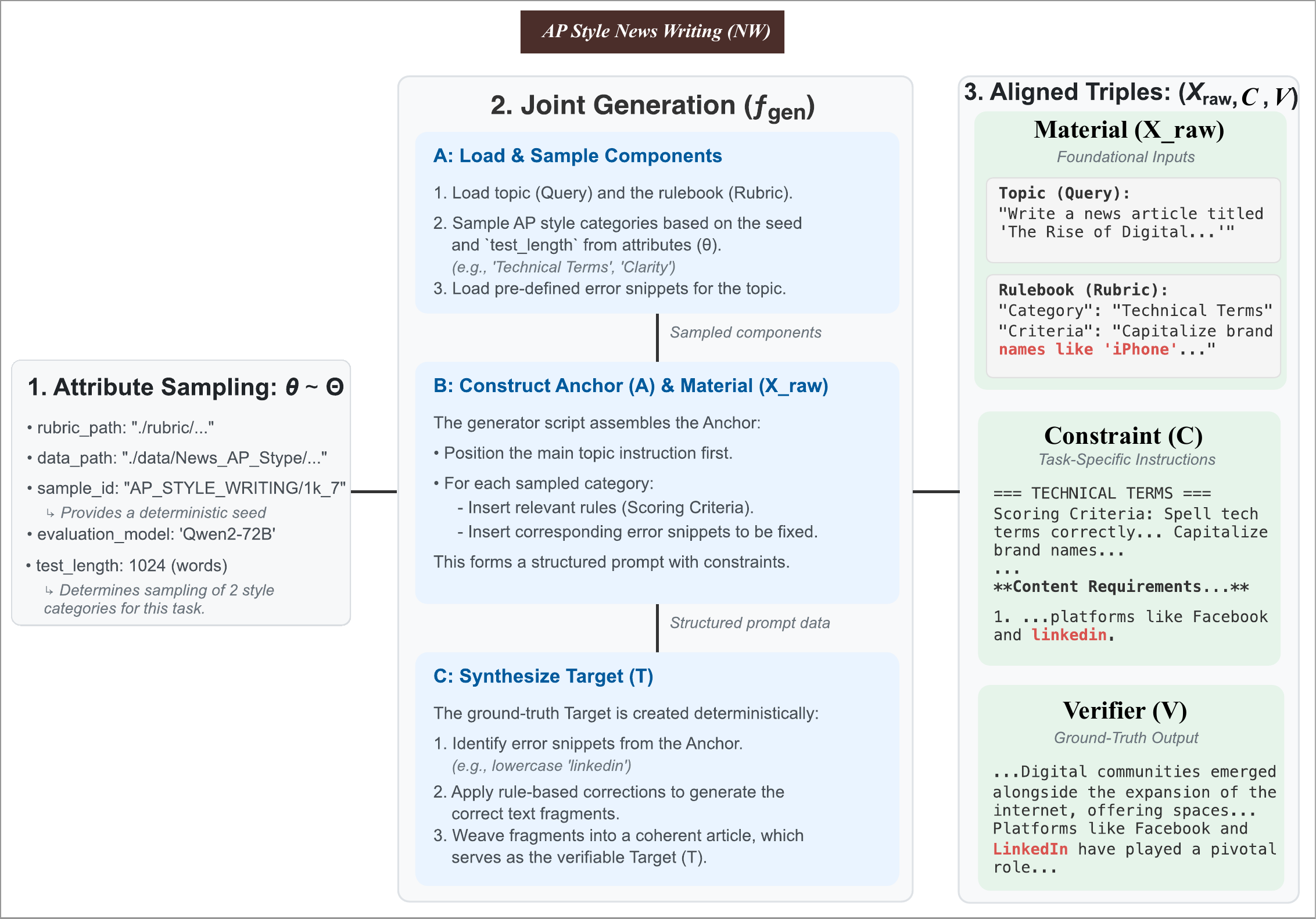}
    \caption{Overview of the data generation pipeline for the AP Style News Writing (NW) task.}
    \label{fig:generator_nw}
\end{figure*}

\begin{figure*}[htbp]
    \centering
    \includegraphics[width=1\textwidth]{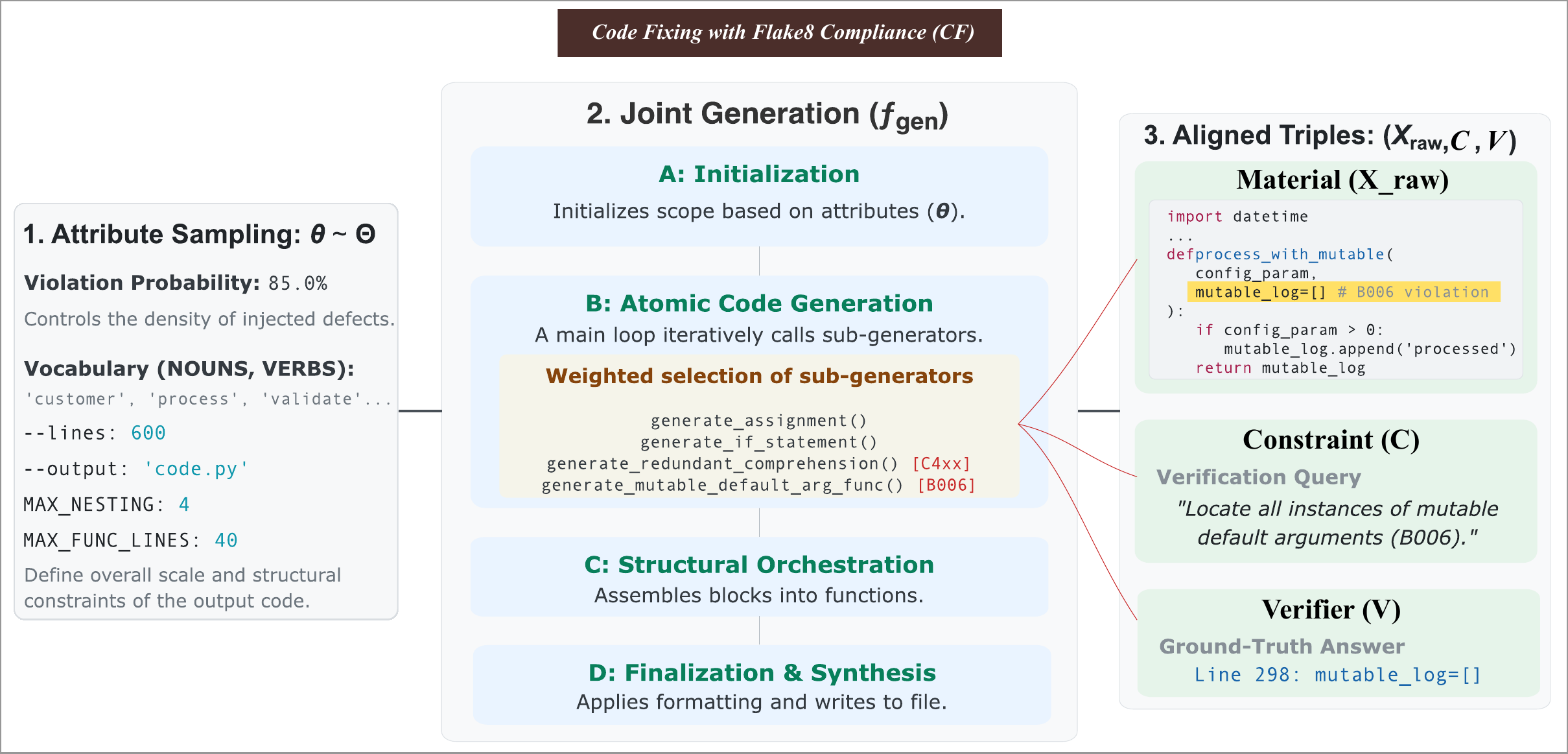}
    \caption{Overview of the data generation pipeline for the Code Fixing (CF) task.}
    \label{fig:generator_cf}
\end{figure*}

\begin{figure*}[htbp]
    \centering
    \includegraphics[width=1\textwidth]{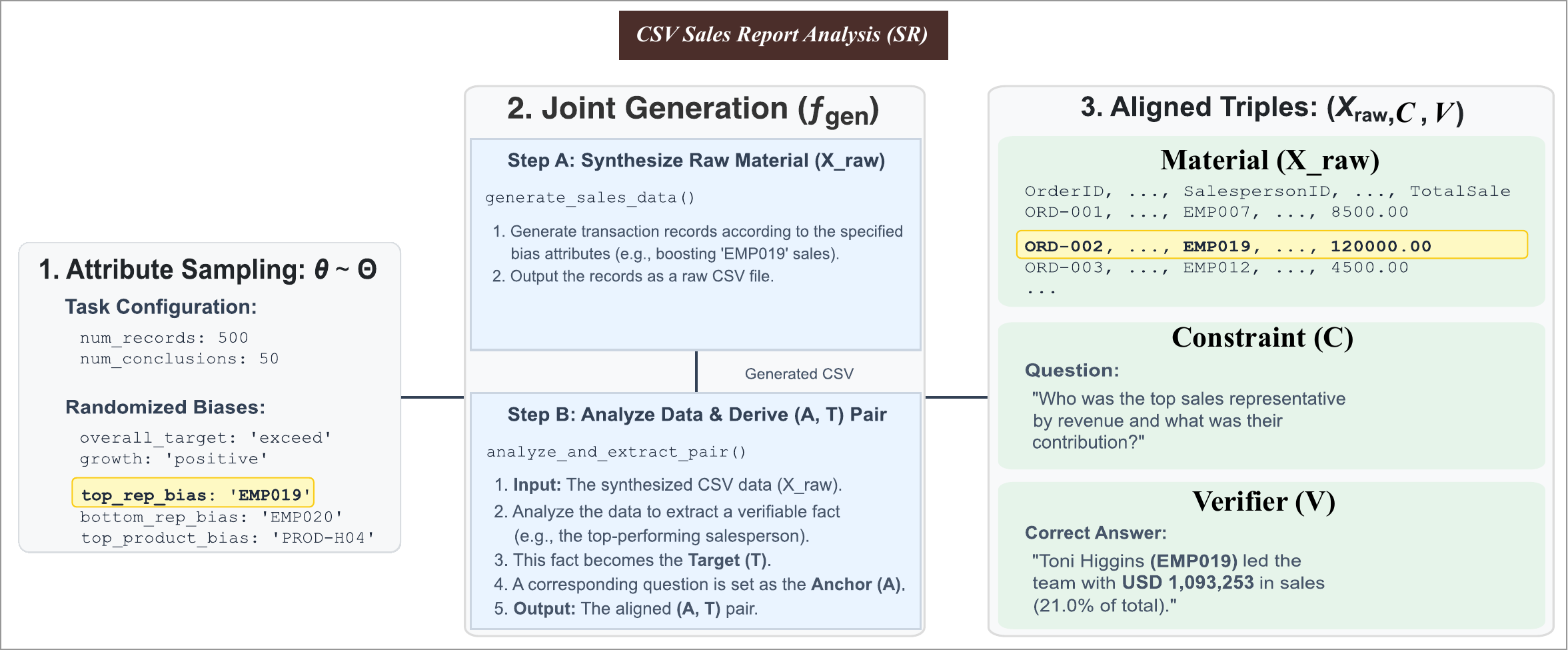}
    \caption{Overview of the data generation pipeline for the CSV Sales Report Analysis (SR) task.}
    \label{fig:generator_sr}
\end{figure*}

\begin{figure*}[htbp]
    \centering
    \includegraphics[width=1\textwidth]{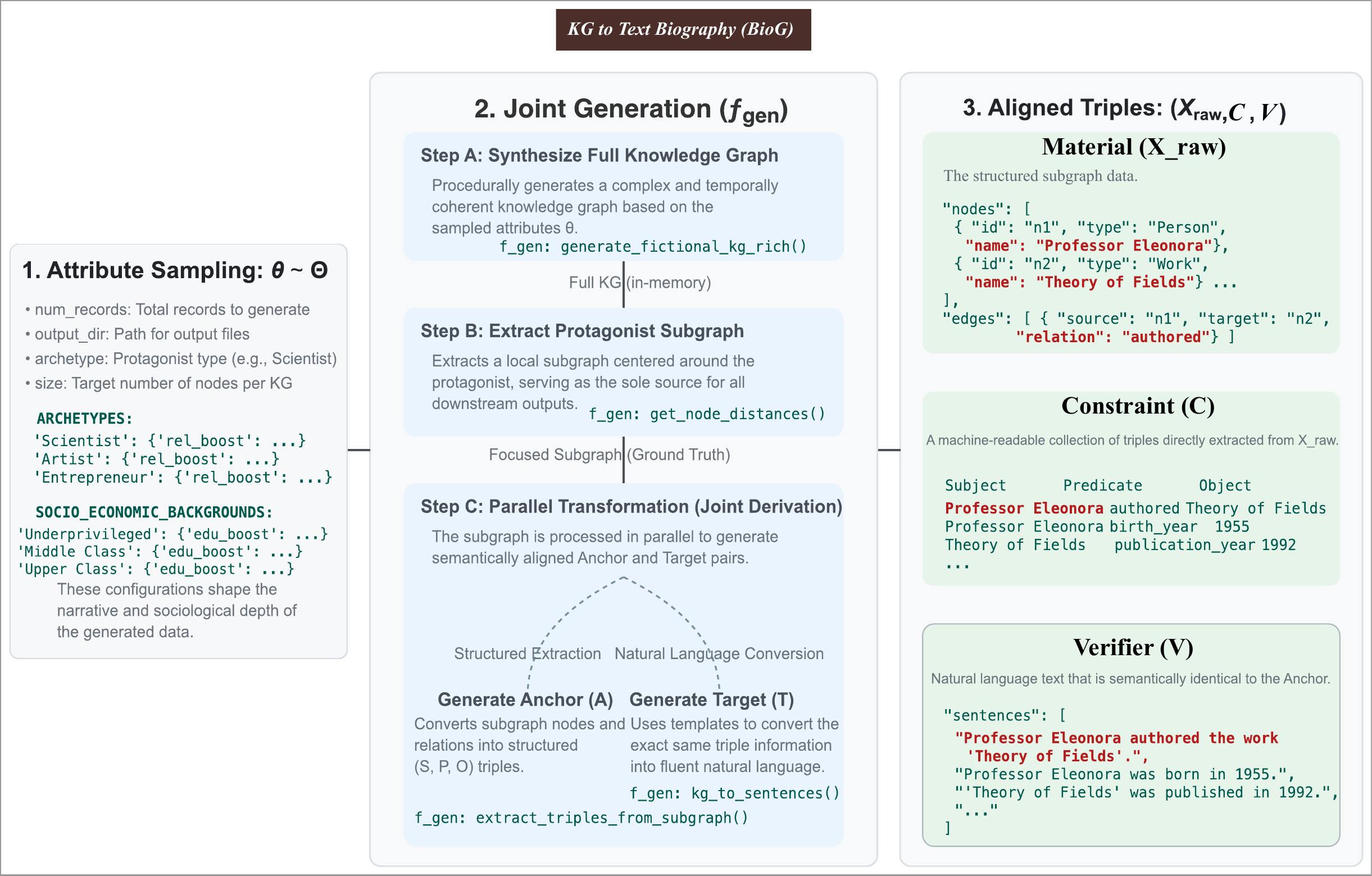}
    \caption{Overview of the data generation pipeline for the KG to Text Biography (BioG) task.}
    \label{fig:generator_biog}
\end{figure*}

\begin{figure*}[htbp]
    \centering
    \includegraphics[width=1\textwidth]{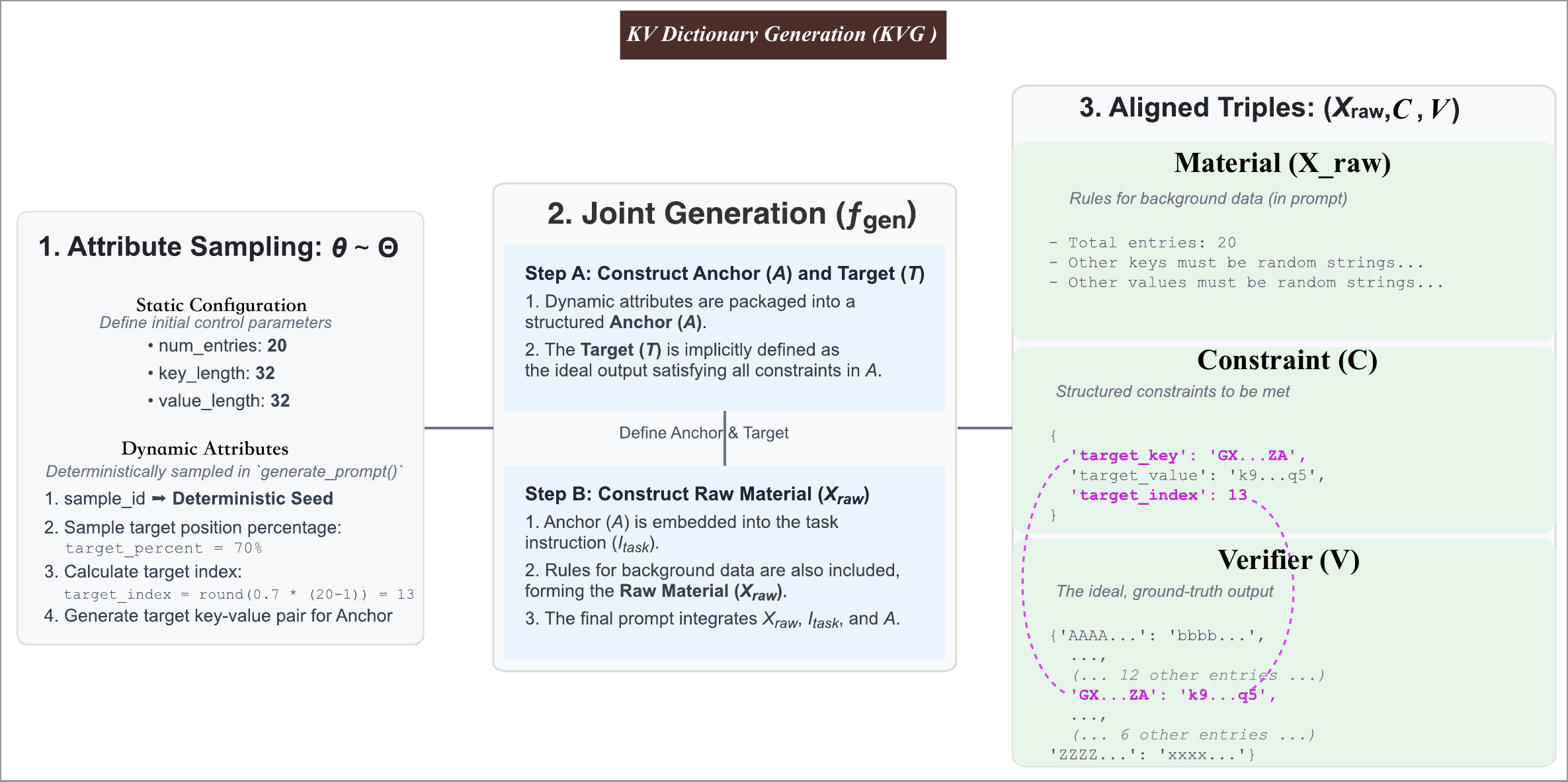}
    \caption{Overview of the data generation pipeline for the KV Dictionary Generation (KVG) task.}
    \label{fig:generator_kvg}
\end{figure*}

\begin{figure*}[htbp]
    \centering
    \includegraphics[width=1\textwidth]{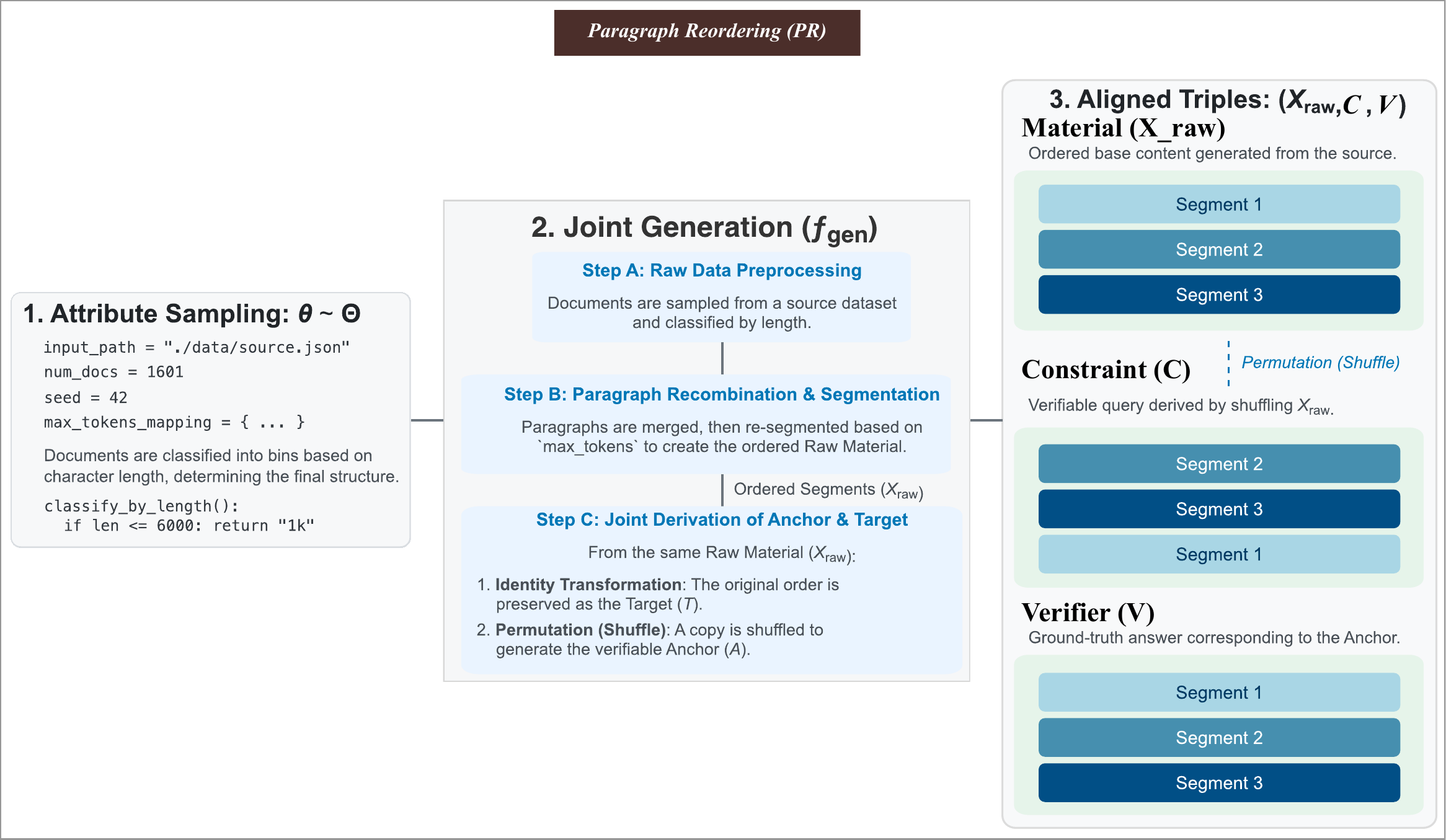}
    \caption{Overview of the data generation pipeline for the Paragraph Reordering (PR) task.}
    \label{fig:generator_pr}
\end{figure*}

\begin{figure*}[htbp]
    \centering
    \includegraphics[width=1\textwidth]{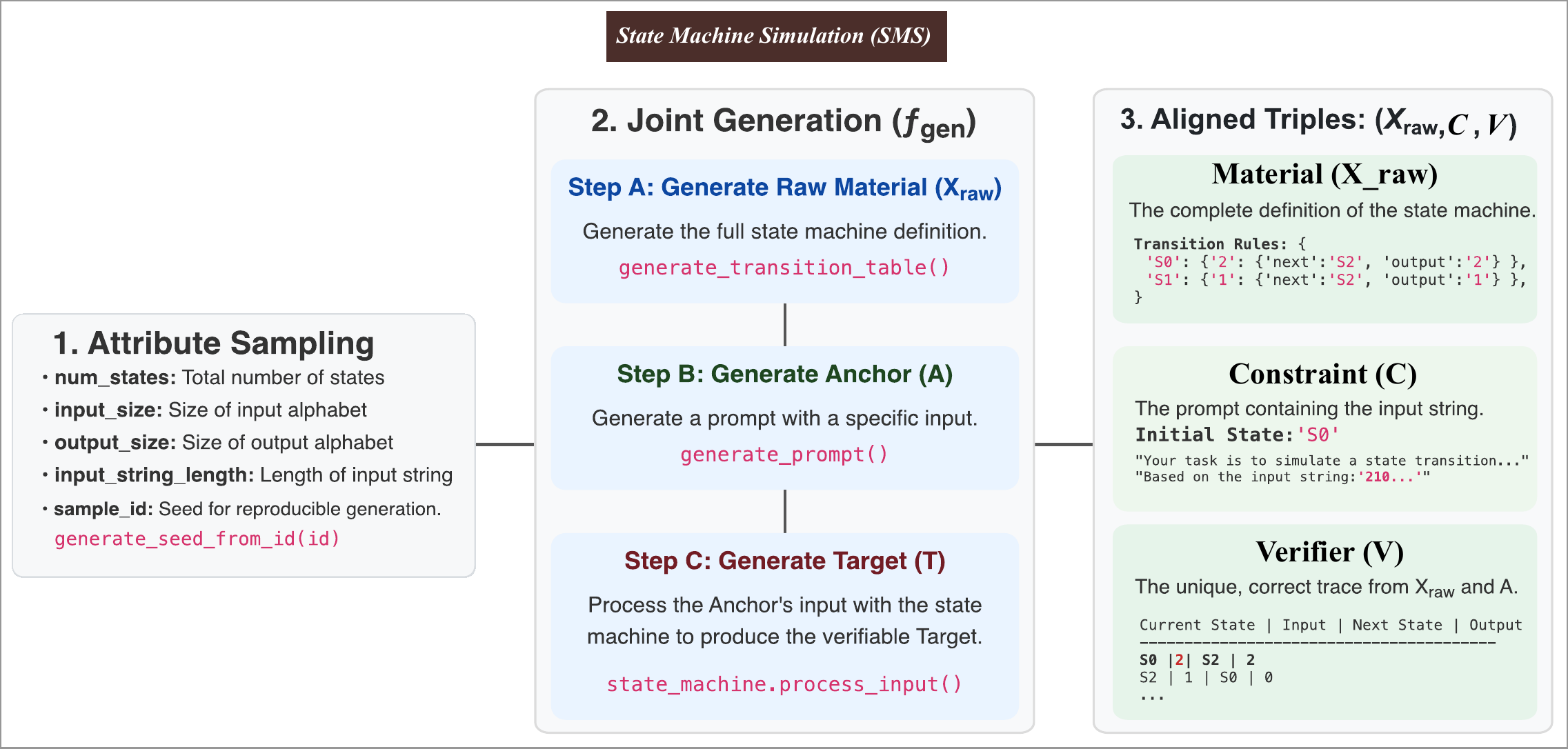}
    \caption{Overview of the data generation pipeline for the State Machine Simulation (SMS) task.}
    \label{fig:generator_sms}
\end{figure*}

To ensure transparency and reproducibility, all test samples in LongWeave are synthetically generated through rule-based pipelines. The generation process for each task, illustrated in \Cref{fig:generator_nw,fig:generator_cf,fig:generator_sr,fig:generator_biog,fig:generator_kvg,fig:generator_pr,fig:generator_sms}, follows a consistent three-stage process. First, \textbf{Attribute Sampling} defines the core parameters and complexity of each task instance. Second, \textbf{Joint Generation} uses these attributes to procedurally create the aligned triad of \texttt{Material} ($X_{\text{raw}}$), \texttt{Constraint} ($C$), and \texttt{Verifier} ($V$). Finally, the third stage in each figure provides a \textbf{concrete example} of the generated \texttt{Material}, \texttt{Constraint}, and \texttt{Verifier}.

\end{document}